\documentclass[final,5p,times,twocolumn,authoryear]{elsarticle}
\usepackage{amssymb}
\usepackage{amsmath}
\usepackage{amsmath,amsfonts}
\usepackage{algorithmic}
\usepackage{algorithm}
\usepackage{array}
\usepackage{float}
\usepackage[
 font=footnotesize
 ]{subfig}
\usepackage{textcomp}
\usepackage{stfloats}
\usepackage{url}
\usepackage{verbatim}
\usepackage{graphicx}
\usepackage{cite}

\usepackage{graphicx}
\usepackage{booktabs}
\usepackage{multirow}
\usepackage[utf8]{inputenc}
\usepackage{pifont}
\usepackage{newunicodechar}
\usepackage{xspace}
\usepackage[pagebackref,breaklinks]{hyperref}
\usepackage{xcolor}

\newcommand{\totalleaks}{2,851\xspace}
\newcommand{\totalnotified}{834\xspace}
\newcommand{\totalprocessed}{25,024\xspace}

\newcommand{\howmanymonthsdeployedN}{11\xspace}
\newcommand{\howmanymonthsdeployedT}{eleven\xspace}

\usepackage{ifthen} 
\usepackage{changepage}

\newboolean{review} 
\setboolean{review}{false} 

\ifthenelse{\boolean{review}}{
\usepackage{ulem} 

\newcommand{\fix}[1]{{\color{blue}#1}}
\newcommand{\fixS}[1]{{\color{blue}\sout{#1}}}

}{
\newcommand{\fix}[1]{#1}
\newcommand{\fixS}[1]{}

}

\journal{Remote Sensing of Environment}
\begin{document}
\begin{frontmatter}

\title{Operational machine learning for remote spectroscopic detection of CH$_{4}$ point sources}

\author[ox,un,jpl]{Vít Růžička\corref{cor1}}
\ead{ruzicka@jpl.nasa.gov}
\author[un]{Gonzalo Mateo-García}
\author[un,upv]{Itziar Irakulis-Loitxate}
\author[un]{Juan Emmanuel Johnson}
\author[un]{Manuel Montesino San Martin}
\author[un]{Anna Allen}
\author[un]{Alma Raunak}
\author[un]{Carol Castaneda}
\author[upv]{Luis Guanter}
\author[jpl]{David R. Thompson}
\cortext[cor1]{Corresponding author}

\affiliation[ox]{organization={University of Oxford},
            city={Oxford},
            country={UK}}

\affiliation[un]{organization={International Methane Emissions Observatory, United Nations Environment Programme (UNEP)},
            }

\affiliation[upv]{organization={Universitat Politècnica de València (UPV)},
            city={València},
            country={Spain}
            }

\affiliation[jpl]{organization={Jet Propulsion Laboratory, California Institute of Technology},
            city={Pasadena},
            state={CA},
            country={USA}
            }

\begin{abstract}
Mitigating anthropogenic methane sources is one \fix{of} the most cost-effective levers to slow down global warming.
While satellite-based imaging spectrometers, such as EMIT, PRISMA, and EnMAP, can detect these point sources, current methane retrieval methods based on matched filters produce a high number of false detections requiring laborious manual verification. 
To address this challenge, we deployed a machine learning system for detecting methane emissions within the Methane Alert and Response System (MARS) of the United Nations Environment Programme's International Methane Emissions Observatory. \fix{This represents the first operational deployment of automated methane point-source detection using spaceborne imaging spectrometers, providing regular global coverage and scalability to future constellations with even higher data volumes.}

\fix{Adapting plume detection for operational deployment required several technical advances.} First, we created \fix{one of} the largest and most diverse \fix{and} global \fix{machine learning ready} datasets \fix{to date} of annotated methane plumes from three imaging spectrometer missions, and quantitatively compared different deep learning model configurations. Second, we extended prior evaluation methodologies from small, tiled datasets to full granules that are more representative of operational use.
This revealed that deep learning models still produce a large number of false detections, a problem we addressed with model ensembling, which reduced false detections by over 74\%. 
Deployed in the MARS pipeline, our system processes scenes and proposes plumes to analysts, accelerating detection and analysis.
During \fix{\howmanymonthsdeployedT}months of operational deployment, \fix{it processed more than 25,000 hyperspectral products facilitating} the verification of \fix{\totalleaks} distinct methane leaks, \fix{which resulted} in \fix{\totalnotified} stakeholder notifications.
We further demonstrate the model's utility in verifying mitigation success through case studies in Libya, Argentina, Oman, and Azerbaijan.
Our work represents a critical step towards a global AI-assisted methane leak detection system, which is required to process the dramatically higher data volumes expected from current and future imaging spectrometers. 
\end{abstract}


\end{frontmatter}


\section{Introduction} 

The reduction of anthropogenic methane emissions represents a critical component of global climate mitigation strategies. As the second second-largestbutor to global warming after CO$_{2}$ \citep{kuylenstierna2021global_UN}, methane's short atmospheric lifetime makes its reducing it a priority lever for achieving the climate targets outlined in the Intergovernmental Panel on Climate Change (IPCC)~\citep{lee2023ipcc}.

A key initiative of the United Nations Environmental Programme (UNEP) International Methane Emissions Observatory (IMEO) is the Methane Alert and Response System (MARS), which detects large point sources of methane emissions using satellite data. MARS utilizes a diverse array of imagery from various public space missions to identify plumes, notify stakeholders, and monitor sources globally over time. This initiative leverages both multispectral and hyperspectral satellite data. Although multispectral sensors offer higher data frequency and longer historical archives, they have limitations over heterogeneous landscapes such as dark, vegetated, or urban areas \citep{varon2020high, gorrono2023understanding}. In contrast, hyperspectral sensors provide the enhanced sensitivity required to detect smaller emissions, even across the most complex surfaces \citep{irakulis2022s2, guanter2021mapping, roger2024high}.

The current landscape of hyperspectral remote sensing missions includes a diverse and growing array of data sources. These include airborne hyperspectral sensors, such as NASA's AVIRIS series and EDF's MethaneAIR; instruments hosted on the International Space Station (ISS), such as NASA's EMIT; and a host of satellite missions, including DLR's EnMAP, ASI's PRISMA, EDF's MethaneSAT, the Chinese Gaofen (GF-5) and Ziyuan (ZY-1) series, and the recently launched Tanager-1 mission from Carbon Mapper and Planet. Furthermore, this collection of sensors is expected to expand significantly, with several next-generation methane-sensitive satellites under development, such as NASA's \fixS{SBG} \fix{upcoming global VSWIR satellite} (estimated to generate 10x more data than EMIT) and Carbon-I, ESA's CHIME, or DLR's CO2Image.

The increasing volume of data from current and forthcoming hyperspectral missions requires the automation of methane emission detection.
Existing detection workflows, which are based on labour-intensive manual review, are inefficient for systematically monitoring large areas and are only suited to monitoring and studying previously identified emission sources. Consequently, a critical need exists for automated tools that can support analysts in monitoring not only known sites but also in discovering new emission events over vast geographical regions. Furthermore, automation can systematize the otherwise arduous and subjective task of delineating the geographic extent of methane plumes, which is a critical step for quantifying emission rates that currently depends on inconsistent human interpretation.

This paper describes a fully operational methane plume detection system integrated into MARS at UNEP's IMEO. By automating the initial screening of hyperspectral satellite data, the models developed in this study facilitate the systematic monitoring of vast areas and the discovery of previously unknown emitters. This capability substantially reduces the manual workload for analysts, allowing a small team to efficiently process a global data stream. In the operational workflow, the system flags candidate plumes for subsequent manual verification, which ensures the correctness and identifies relevant stakeholders for notification \citep{eye_methane_2024}. 
The integration of this model into the MARS pipeline enhances operational efficiency by enabling timely detection of emission events for rapid stakeholder engagement and by scaling monitoring capabilities to support a more comprehensive and systematic analysis of methane dynamics and regulatory accountability. 
\fix{The system processed \fix{\totalprocessed} scenes, resulting in \fix{\totalleaks} verified methane leaks, and \fix{\totalnotified} stakeholder notifications over the first \fix{\howmanymonthsdeployedT} months of operation of the model.  These results show the model's utility not only for initial detection but also for the verifying successful mitigation, substantiated by case-studies in Libya, Argentina, Oman, and Azerbaijan.} 

\fix{Apart from the deployment outcomes, we also describe several distinct technical advances that were required to scale the machine learning system for this task: (i) the curation and public release of the largest and most diverse global dataset of manually annotated methane plumes, compiled from three distinct hyperspectral missions; (ii) a rigorous evaluation of various model configurations and a comprehensive performance analysis on full satellite granules, culminating in an ensemble approach that reduces false alerts by over 74\% compared to the prior state-of-the-art \citep{STARCOP}; and (iii) the demonstration of successful zero-shot generalisation across sensors, with models trained on EMIT data evaluated on PRISMA and EnMAP, whose performance is further enhanced with fine-tuning.} 

The remainder of the paper is organized as follows: Section~\ref{sec:background} reviews the literature on methane detection and prior machine learning approaches. Section~\ref{sec:data} describes the datasets from the EMIT, PRISMA, and EnMAP missions. Section~\ref{sec:methods} details our machine learning methodology and Section~\ref{sec:experiments} outlines our experimental setup for both single-sensor training and cross-sensor adaptation. Section~\ref{sec:results} presents our findings on model performance, false positive reduction, operational deployment results, and mitigation case-studies. Finally, Section~\ref{sec:discussion} offers concluding remarks and discusses future work.

\section{Background literature}\label{sec:background}

Here, we briefly review prior work using machine learning models for methane detection. For a more in-depth review, we refer the reader to surveys such as \citep{tiemann2024machine} or the background chapters of \citep{ruuvzivcka2025intelligent}.

Early scientific papers of \citep{frankenberg2016airborne, thompson2016space, duren2019california, irakulis2021prisma, irakulis2022s2} systematically explore large selected geographical regions and demonstrate that methane leaks can be tracked with both multispectral (MSI) \citep{varon2020high, sanchez2021mapping, irakulis2022s2} and hyperspectral (HSI) \citep{duren2019california, irakulis2021prisma, guanter2021mapping, roger2024high} instruments. These articles measure methane enhancement using band ratios in MSI data \citep{varon2020high} and matched-filter approaches in HSI data \citep{thompson2015real}, but the identification of discrete point sources remains largely manual.

Soon after, the first automated approaches leveraging machine learning models trained on either raw \citep{joyce2023using} or interpreted satellite data followed \citep{jongaramrungruang2022methanet, groshenry2022detecting}. As highlighted in the opinion piece by \citep{thompson_ml_promise_2021}, research on developing machine learning models for hyperspectral data has been limited by the availability of large, annotated datasets. As such, these early works either considered only a relatively small number of real methane leak events or eddepended on the simulations of synthetic data. NASA presented some of the first large, annotated datasets in the work of \citep{cusworth2021intermittency} using an aerial campaign of AVIRIS-NG over the Permian Basin region in the US. In the works of \citep{STARCOP, kumar2023methanemapper}, this data has enabled training more complex machine learning models on real methane leak events.

Furthermore, \citep{STARCOP} has cleared and refined the ground-truth labels for this dataset and released the data in a machine-learning-ready format. They also showcased some early examples of generalisation across sensors, by detecting events in data from the spaceborne instrument EMIT. The recent work of \citep{mancoridis2025multi} further explores cross-sensor adaptation using generative machine learning models.

\fix{It is also worth noting the work that explores different methods for computing methane enhancement products.}
The most widely used approaches derive from the matched filter method \citep{thompson2015real}, for example, its iterative variant, Mag1c \citep{foote_fast_2020_mag1c}. More recently, variants using extended spectral ranges of imaging spectroscopy data were proposed as the wide-window matched filter in \citep{roger2023wide_MF}. A similar approach is also noted in \citep{bue2025towards} as the full VSWIR variant of the column-wise matched filter (full CMF). However, we note that most of these works do not quantitatively compare these methane enhancement products across large datasets.

Additionally, some very recent research, such as \citep{HyperspectralViTs}, explores end-to-end processing of imaging spectroscopy data without computing the matched-filter product (using radiance cubes as input instead). However, these approaches require adapting traditional machine learning architectures to handle the large number of input bands\fix{, and do not easily generalize across sensors}.

Only a a few recent works consider real-world operational settings and the potential deployment of machine learning models. The work of \citep{vaughan2024ai} demonstrates the detection of methane leaks at selected monitoring locations using multispectral Sentinel-2 and Landsat imagery, with a refined version of the models proposed in \citep{vaughan2023ch4net}. These models are currently used in UNEP’s operational methane plume detection pipeline for processing multispectral data. The recent preprint \citep{bue2025towards}, developed around the same time as our paper, considers using machine learning models as part of the detection pipeline at NASA. At the time of this writing, these models are not yet deployed.

Finally, we note that none of the prior published works have used a dataset of methane leak events as large as the one we are releasing in this paper (including data from 3 different sensors). Our comparative study of different methane enhancement products is also novel at this scale. As far as we know, our machine learning models are also the first operationally deployed and daily-used models for detecting methane leaks in hyperspectral data, enabling event discovery in novel regions.

In this paper, we present datasets created from three hyperspectral sensors. We compute several methane enhancement products from this data and use them to train machine learning models, as described in recent research \citep{STARCOP}. We explore several additions to these models, such as using wind and location as additional inputs. Finally, we report results on realistic test sets created to mirror the production environment, namely, in using the entire captured image (the full tile). We identify shortcomings of existing models and explore the usage of model ensembles to address these limitations.

\section{Data}\label{sec:data}

\begin{figure}[!h]
  \centering
  \subfloat[a][EMIT]{\includegraphics[trim={0 0 0 0cm},clip,width=1.0\linewidth]{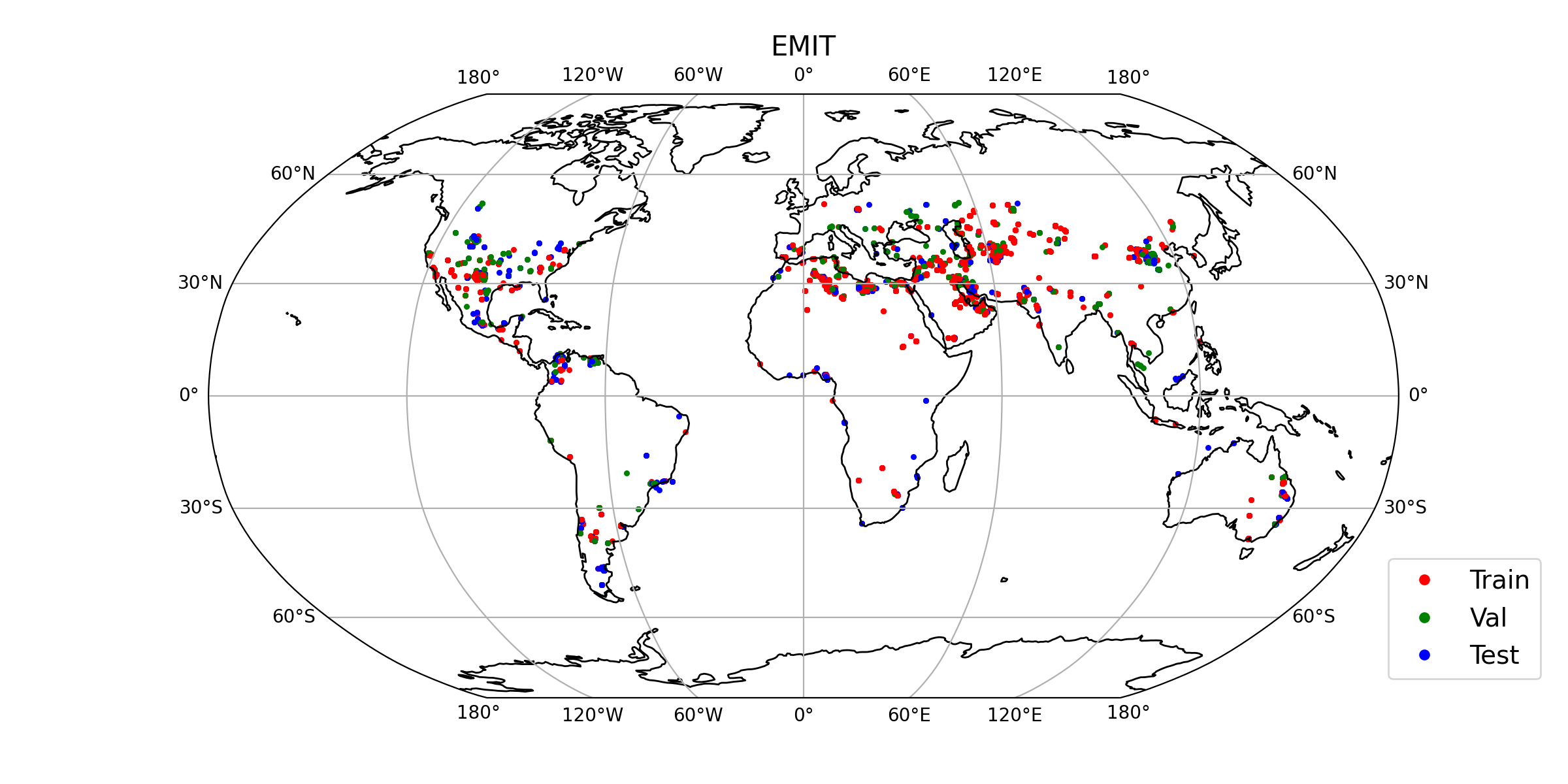}\label{fig:map_a}} \\
  \subfloat[b][PRISMA]{\includegraphics[trim={0 0 0 0cm},clip,width=0.9\linewidth]{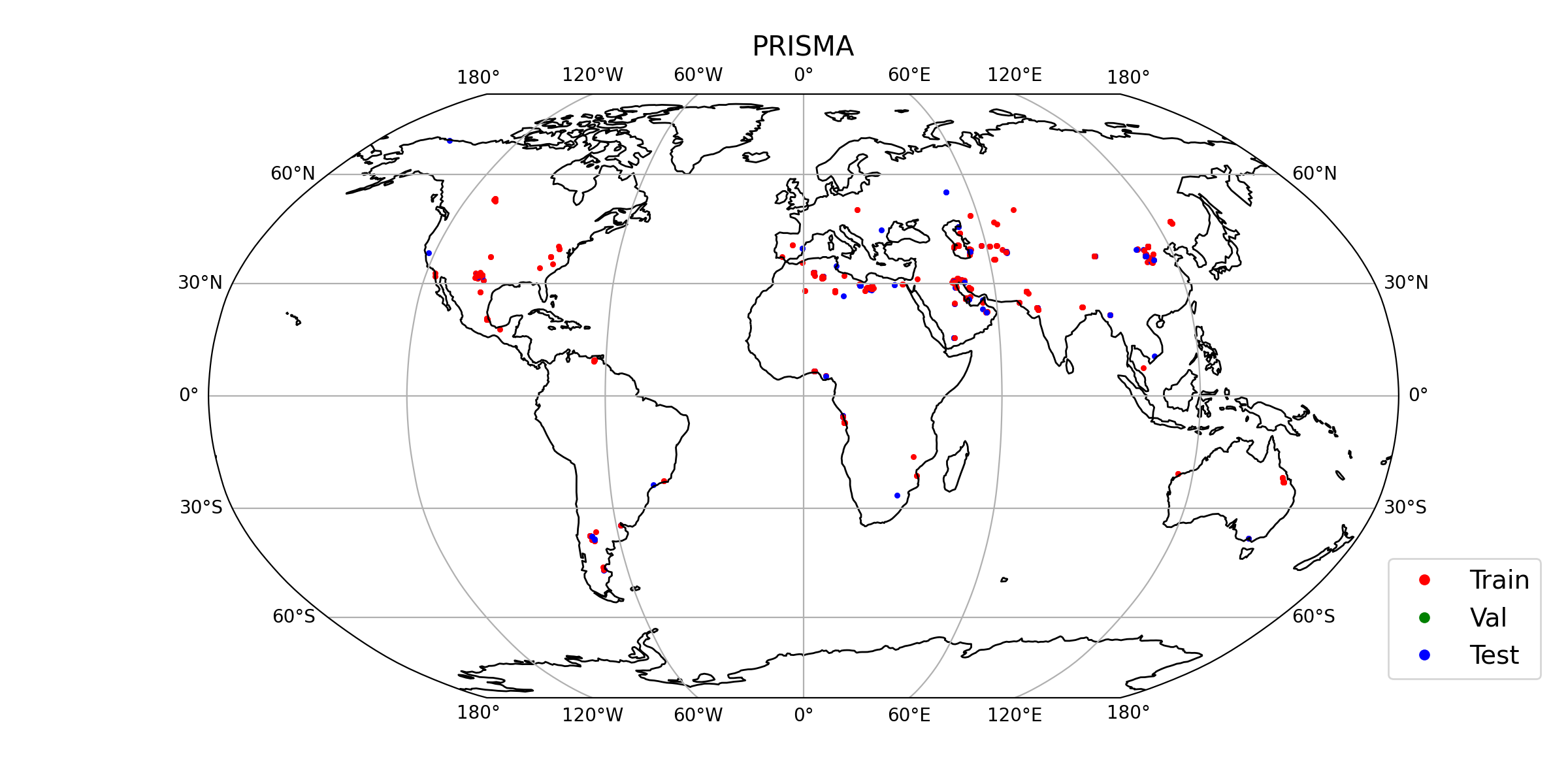}\label{fig:map_b}} \\
  \subfloat[c][EnMAP]{\includegraphics[trim={0 0 0 0cm},clip,width=1.0\linewidth]{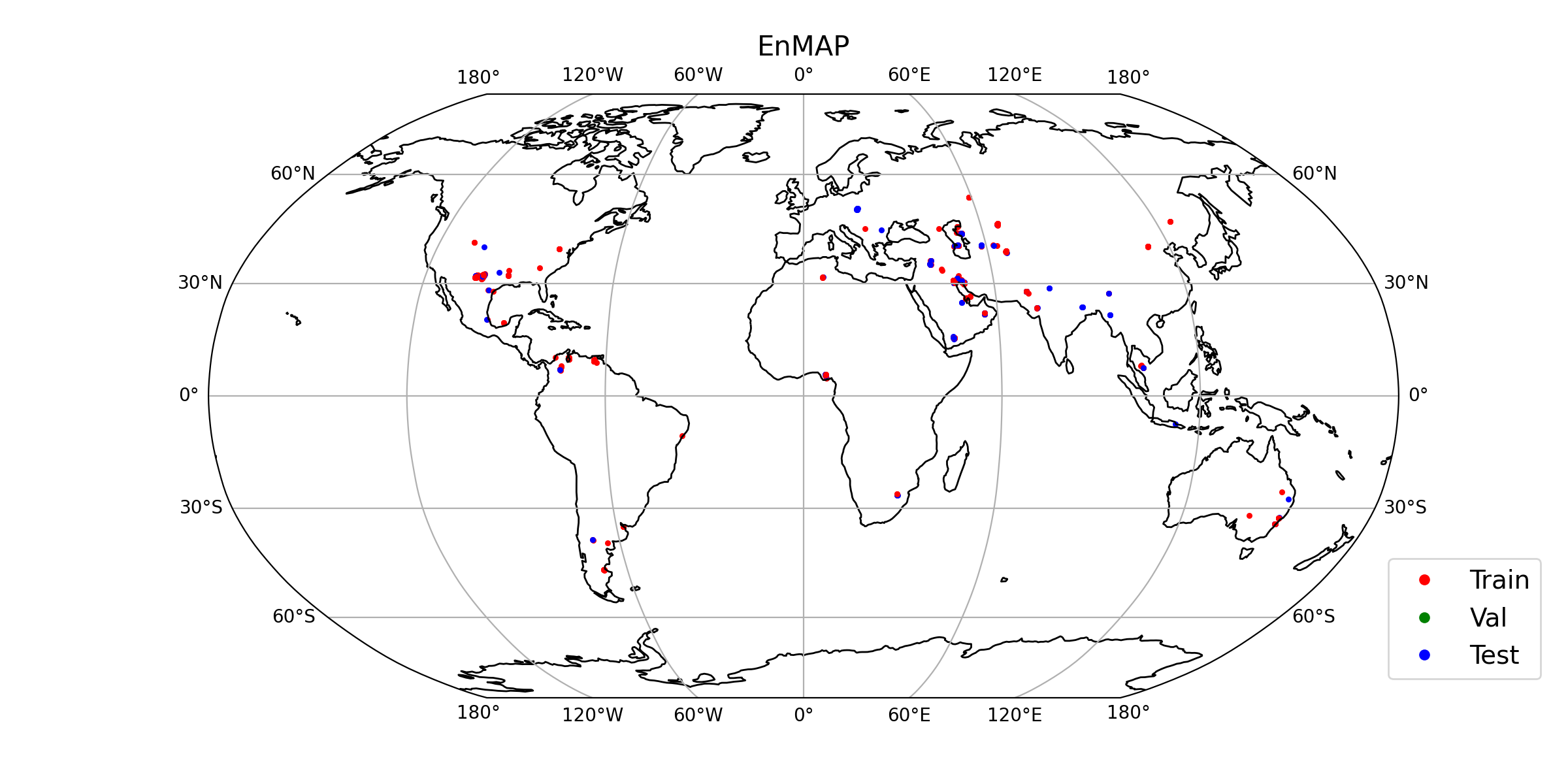}\label{fig:map_c}}
  \caption{Spatial distribution of samples in the created datasets and their division into train (red), validation (green, only for EMIT), and test (blue) subsets.  \label{fig:dataset_maps}}
\end{figure}

The datasets used in this paper originate from the operational archive of UNEP IMEO's Methane Alert and Response System (MARS), which contains validated emission events from September 2022 to March 2025; an extended version of this data is detailed in appendix C. in Section \ref{sup_extended_dataset}. Within this framework, a team of analysts identified numerous methane leaks across various sectors and geographic areas, annotating each with high-quality semantic segmentation outlines. A key feature of this dataset is its rigorous validation: every detection was verified by at least two independent analysts, with a third confirmation required for any plume notified to stakeholders. \fix{ Similarly, plume-free scenes were independently reviewed to confirm the absence of methane emissions}. 
This work leverages these verified detections from three hyperspectral sensors: EMIT \citep{green2022_EMIT}, PRISMA \citep{cogliati2021prisma}, and EnMAP \citep{EnMAPmission}. The global coverage of the resulting datasets is visualized in Figure~\ref{fig:dataset_maps}.

The annotation process starts with the analysts primarily looking at the matched filter products (by default WMF, but they can also choose to display other variants). They also see several overlaid vector layers showing known infrastructures, a high-resolution base map (both of these can be used to identify possible source structures on the ground) the RGB product from the sensor itself (useful in cases when the base map may be outdated), and additional metadata such as the wind direction and speed. Finally, and this is important in cases of weak or uncertain methane plumes, they have access to previous (and future) observations of the same scene (with several sensors, including EMIT, PRISMA, and EnMAP). This is useful for distinguishing a methane enhancement from false enhancements caused by surface structures or mineralogical features. For plume-free image sources, we used locations that have been verified not to have any methane leak events.

Importantly, the datasets we are releasing in our work are made to be machine learning ready - this means that we attempt to lower the entry barriers for machine learning researchers using our data. While detection catalogues such as the UNEP IMEO's Eye on Methane platform (\url{https://methanedata.unep.org/map}), Carbon Mapper's data portal (\url{https://data.carbonmapper.org}), or the product published in NASA's Visions portal (\url{https://earth.jpl.nasa.gov/emit/data/data-portal/Greenhouse-Gases/}) and in \citep{green2023emit_plumes_data_portal_product}, provide information about methane emission events, they are not intended for machine learning purposes. Hence, we provide a machine-learning ready dataset, including the following features:
(i)~standardised input and output products, (ii)~curated non-overlapping data splits, (iii)~positive and negative samples, and (iv)~dedicated full-granule test sets.
The steps described in our paper can be used to construct similar datasets from these other sources of data.

\subsection{EMIT dataset}
\label{sec:data-emit}

The EMIT dataset currently represents the largest released \fix{machine learning-ready} dataset of methane leak events from a hyperspectral sensor with global coverage \fix{so far}. EMIT is an imaging spectrometer deployed on the ISS that has provided near-global coverage of Earth since its launch in July 2022. The mission has been extended at least until 2026. EMIT captures images with a ground resolution of 60 m, with spectral bands between 381 nm and 2493 nm, and a spectral resolution of approximately 7.5 nm. The sensor produces images of an area of about 76.8 km x 74.52 km. We refer to these as {\it full tile} or {\it full granule} captures.

\begin{figure}[!h]
    \centering
    \includegraphics[width=1.0\linewidth]{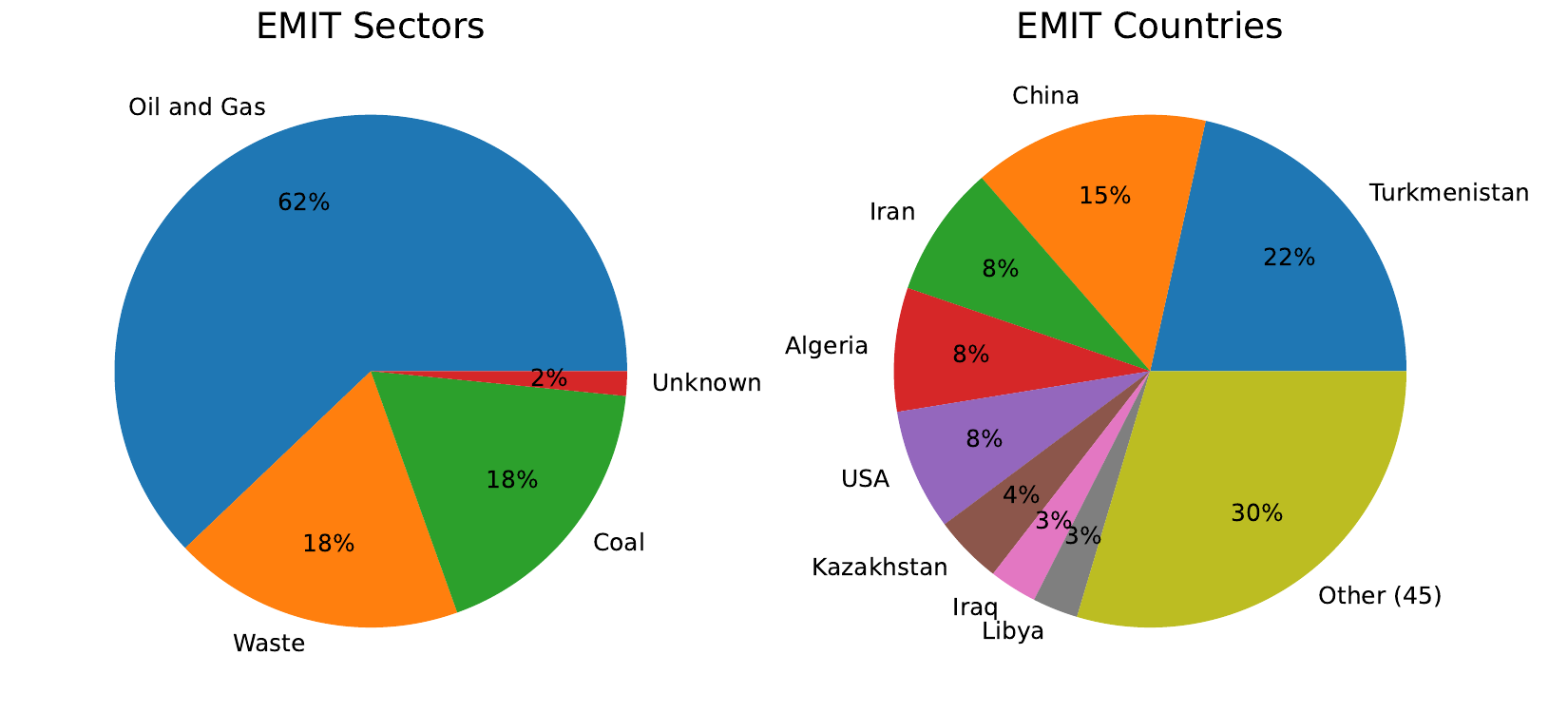}
    \caption{Stratification of EMIT data by sectors and countries.}
    \label{fig:emit_stratification}
\vspace{-2mm}
\end{figure}

We create our dataset from 3806 methane leak events detected by the IMEO team. Figure \ref{fig:emit_stratification} details the stratification of these events into sectors and countries. An additional 5965 locations that have been validated as not containing any plume leaks are included in this dataset. 
In total, the dataset uses 3916 unique EMIT captures, 1528 of which contain plumes.

\begin{figure}[!h]
    \centering
    \includegraphics[width=1.0\linewidth]{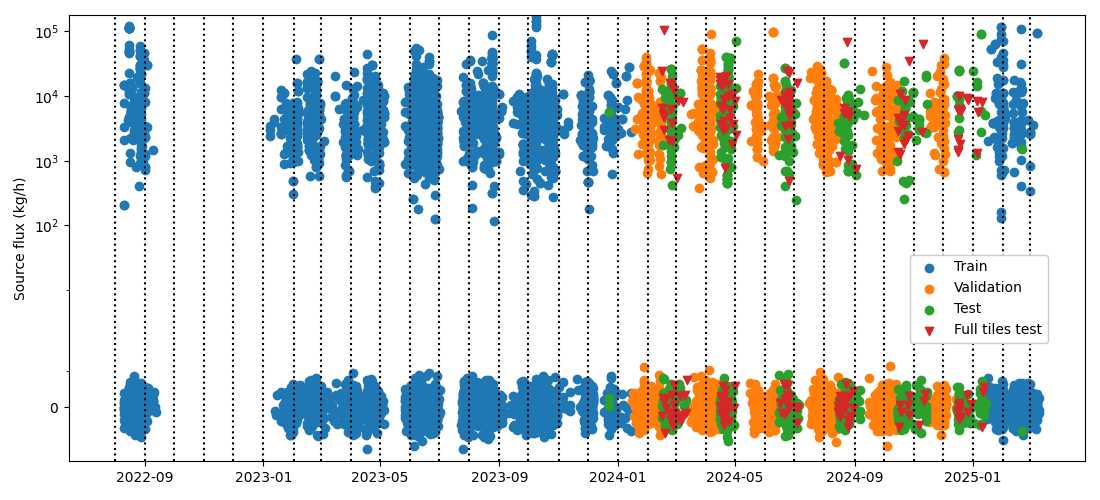}
    \caption{Temporal distribution of samples in the EMIT dataset. Training, validation, and test subsets are also shown; full tiles are downloaded for selected samples from the test subset. Note that for no-plume tiles (with flux rate equal to 0), we jitter these points around the x-axis for better visualisation. }
    \label{fig:emit_temporal_split}
\vspace{-2mm}
\end{figure}

These sources add up to more than 7.8 Tb of data, so extracting small tiles from these full scenes is necessary to reduce the size. Specifically, we extract smaller subregions of 256x256px (about 15.36 x 15.36 km)\fixS{ as geotiff files} to reduce data storage requirements. This data is then kept in the GeoTIFF format. These tiles are centered around methane leak events, so further tiling is suggested to break this regularity. It is important to bookkeep which tiles overlap so that they can be kept in the same validation splits. To this end, we split tiles into train, validation, and test subsets using selected date ranges, which ensures that different splits do not use the same source scene. These splits are shown in Figure \ref{fig:emit_temporal_split}. For training, they are further tiled and augmented to match the models' input resolution. A selected sample of data from the test split, 105 plume events and 95 no-plume images, is also downloaded as full EMIT tiles.

\subsection{PRISMA and EnMAP datasets}

We create two other datasets from the hyperspectral sensors PRISMA and EnMAP. 
In contrast to the larger EMIT dataset, we choose to use spatial rather than temporal data splits.

As these satellites operate in on-demand mode (whereas EMIT captures all ground-based overpassed locations), the number of methane leak events is smaller - this highlights the fact that manual scheduling of satellite targets misses many events that may only be found later in the globally collected data archive.
This allows us to explore a general multi-sensor scenario in which we have a very large dataset from a single sensor and smaller datasets from other sensors. This will be a typical situation for newly deployed sensors that have observed only a few methane leak events.

The PRISMA dataset consists of 466 plume events and 664 selected no-plume locations. The PRISMA sensor observes spectral bands between 400 and 2505 nm, with spectral resolution of approximately 11-13 nm and ground resolution of 30 m. We extract 256x256 px tiles and split them into spatially non-overlapping train and test subsets. For the entire test set, we downloaded the full sensor data of 115 images with a total of 63 methane leak events and 52 background scenes.

The EnMAP dataset is the smallest, consisting of 239 plume events and 320 no-plume locations. The EnMAP sensor has spectral bands between 420 and 2450 nm, with a spectral resolution of 6.5 nm and a ground resolution of 30 m. As for the other datasets, we extract 256x256 px tiles and split them into spatially non-overlapping train and test subsets. The entire test set includes 83 full-tile images, which contain 54 methane-leak events and 29 background scenes.

For all datasets, we also include the wind information from the U$_{10}$ NASA GEOS-FP reanalysis product of \citep{molod2012geos_windU10}.

Detailed statistics about the datasets and the created data splits are in Table \ref{tab:dataset_statistics}.

\begin{table}[h]
\caption{Overview of datasets created in this paper. Positive and negative tiles refer to the 256x256 px tiles extracted from the full images - \fix{we mark tiles as positives, as long as there is a labelled plume inside the area. For full tile test sets, we keep the scenes in their original resolution.} Sites refer to unique locations. Meanwhile, the number of original granules refers to the total number of separate satellite captures and can be used to compare the scale of these datasets with other sources. We note that during training, the train splits are further tiled into smaller resolution (128x128 px with 64 px overlap and several augmentations). }
\label{tab:dataset_statistics}
\centering
\scalebox{1.0}{
\begin{tabular}{@{}lccc@{}}
\toprule
\multicolumn{4}{l}{\textbf{MARS-EMIT dataset}}                                                       \\ \midrule
\textbf{Split} & \textbf{Pos. : Neg. tiles} & \textbf{Sites}       & \textbf{Original granules} \\
Train          & \fix{2796 : 2950}              & 1590                 & 2250                       \\
Val            & \fix{957 : 1359}               & 1244                 & 954                        \\
Test           & \fix{824 : 884}                & 1102                 & 712                        \\ \midrule
Full tile test & 200                 & -                    & 200                        \\
               & \multicolumn{1}{l}{}            & \multicolumn{1}{l}{} & \multicolumn{1}{l}{}       \\ \midrule
\multicolumn{4}{l}{\textbf{MARS-PRISMA dataset}}                                                     \\ \midrule
\textbf{Split} & \textbf{Pos. : Neg. tiles} & \textbf{Sites}       & \textbf{Original granules} \\
Train          & \fix{381 : 523}                 & 467                  & 342                        \\
Test           & \fix{115 : 111}                 & 192                  & 115                        \\ \midrule
Full tile test & 115                   & -                    & 115                        \\
               & \multicolumn{1}{l}{}            & \multicolumn{1}{l}{} & \multicolumn{1}{l}{}       \\ \midrule
\multicolumn{4}{l}{\textbf{MARS-EnMAP dataset}}                                                      \\ \midrule
\textbf{Split} & \textbf{Pos. : Neg. tiles} & \textbf{Sites}       & \textbf{Original granules} \\
Train          & \fix{172 : 219}                 & 293                  & 156                        \\
Test           & \fix{81 : 87}                   & 151                  & 83                         \\ \midrule
Full tile test & 83                    & -                    & 83                         \\ \bottomrule
\end{tabular}
}
\end{table}

We split each sensor's data into non-overlapping subsets. For our largest dataset of EMIT images, we create three splits (train, validation, and test) using temporally non-overlapping date ranges. Specifically, we reserve all data from 2024 for the test and validation datasets. This allows us to use any images collected before and after for training (data starting from September 2022).  Thus, any future data collected by the EMIT sensor can be added to the training dataset for later iterations of these models.
Note that the ISS-hosted sensor EMIT exhibits different coverage in summer and winter months.
To provide representative subsets in both test and validation data, we divide data by the 15th of each month and assign data ranges beginning with odd months for the validation dataset (e.g.: 15th Jan to 15th Feb, not including the last day) and even months to the test dataset (e.g., 15th Feb to 15th Mar, not including the last day). The last date range (15th Dec 2024 to 15th Jan 2025) is kept in the test dataset. This split is visualised in Figure \ref{fig:emit_temporal_split} and can be exactly reproduced using our released code.

For the PRISMA and EnMAP datasets, we instead use non-overlapping splits spatially. This decision was made because the datasets are smaller, and a temporal data split might introduce unwanted overfitting in some locations. These locations are shown on Figure \ref{fig:dataset_maps}.

\subsection{Methane enhancement products}

An established practice for the detecting methane leak events in hyperspectral data uses so-called methane enhancement products (generally, estimates of methane concentration above background levels). These are classical target detection products computed from the source hyperspectral data. One such family of methods uses the matched filter (MF) operator. Matched filters are used to search for target spectral signatures in hyperspectral data. In the context of methane leak detection, we use a methane signature calculated from HITRAN absorption coefficients \citep{BROWN2013201_HITRAN}.

Matched filter products therefore reduce the hyperspectral datacube into a single band of information representing the enhancement of methane above the background level.  These enhancement maps can then be used by trained specialists or by machine learning models. These products include several error sources, such as instrument noise, as well as artifacts from surface materials that are spectrally similar to methane. These so-called confounders typically include artificial structures such as roads and solar panels, as well as natural features such as mountain ranges and desert dunes.

Several variants of matched-filter products were proposed to reduce these confounding effects. The algorithm called Mag1c \citep{foote2020fast_mag1c} proposes iterative recomputation of matched filters and parameter adjustment for background estimation. The recently proposed wide-window matched filter (WMF) \citep{roger2023wide_MF} instead uses a wider spectral range of the source hyperspectral data to better represent the scene background. 
More concretely, we use the product referred to as MF SWIR by \citep{roger2023wide_MF} computed from the following extended spectral ranges in EMIT data: 976.9nm - 1260nm and the full 1330nm - 2441.1nm.

\begin{figure*}[!h]
    \centering
    \includegraphics[width=1.0\linewidth]{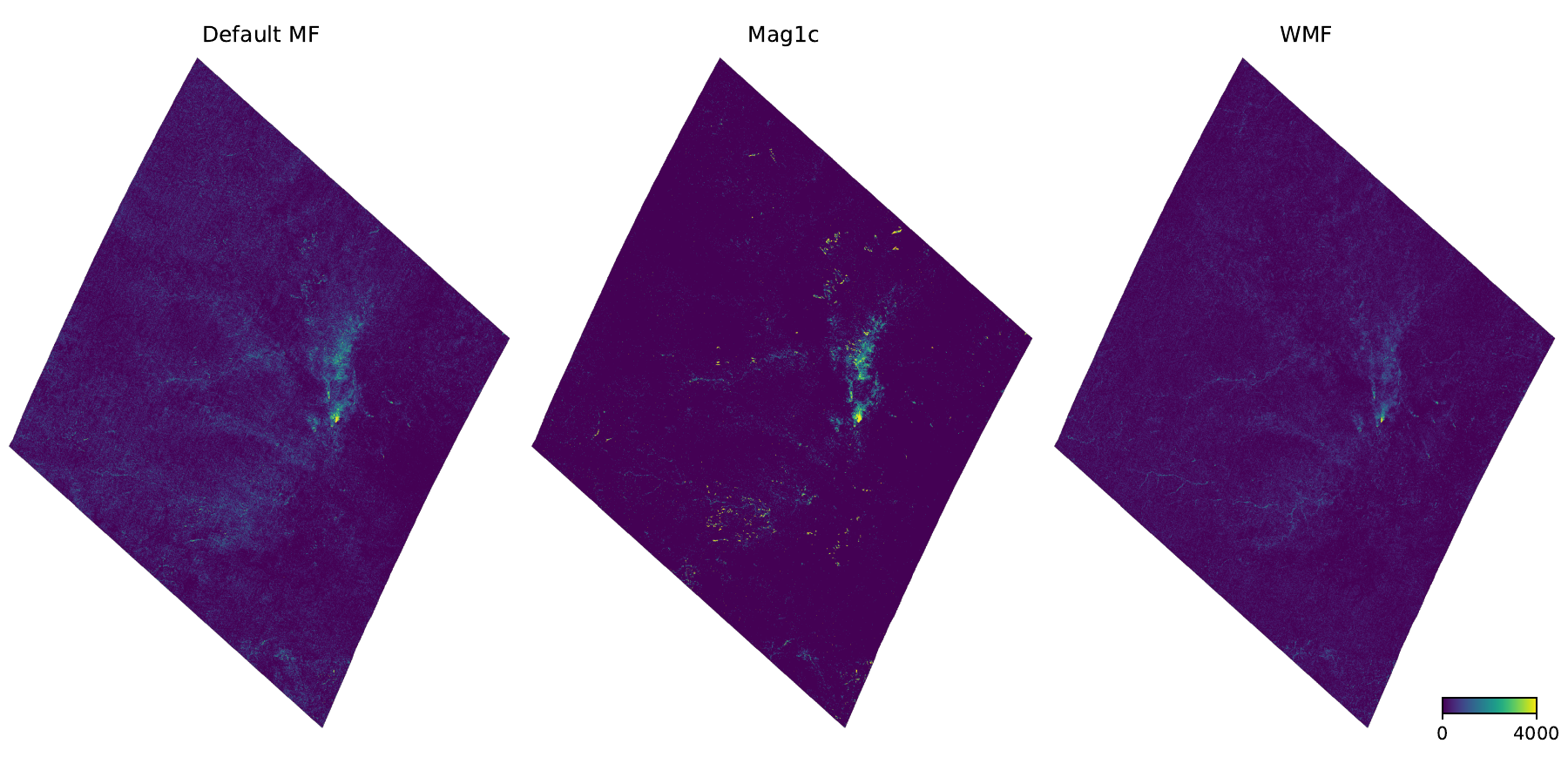}
    \caption{Comparison of the explored methane enhancement products computed for a sample scene from the EMIT full tile test split. For easier comparison, all matched filter products were scaled to the same visualisation range (0 to 4000 ppm$\times$m units corresponding to the mixing ratio length). While some artifacts replicate the structure of the real scene (e.g., rivers and mountain ranges) across all products, the Mag1c product seems to have the most prominent confounders.}
    \label{fig:products_vs}
\vspace{-2mm}
\end{figure*}

Figure \ref{fig:products_vs} shows an example visualisation of the three methane enhancement products. In our work, we compute all matched-filter products for all samples in the datasets considered. This enables us to compare methods using any of these classical products.

\section{Methods}\label{sec:methods}

\subsection{Machine learning models}
\label{sec:methods-ml}

\begin{figure}[!h]
    \centering
    \includegraphics[width=1.0\linewidth]{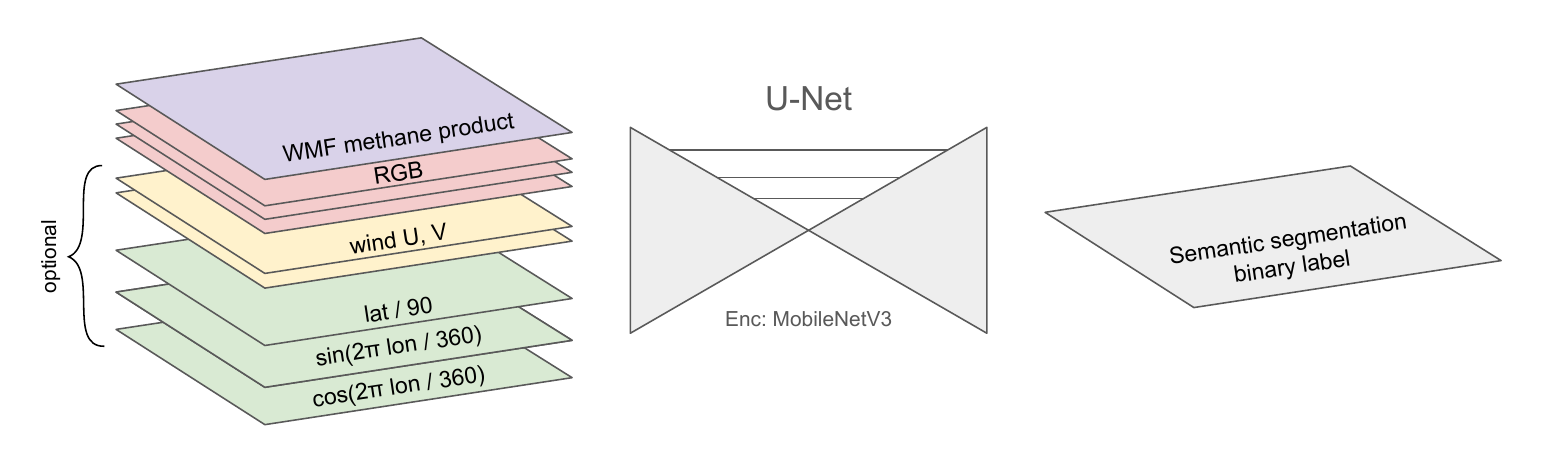}
    \caption{HyperMARS model illustration. We use the U-Net model architecture with the MobileNetV3 encoder and a variety of input features, further described in the text. Depending on the used configuration of input products, the model has around 6.69M trainable parameters. \fix{Note that for operational deployment, we chose to use the simple variant using only WMF and RGB bands in an ensemble of models (the other bands are marked as optional).}}
    \label{fig:model_illustration}
\vspace{-2mm}
\end{figure}

In this paper, we use machine learning architectures based on the HyperSTARCOP models proposed by \citep{STARCOP}.  However, we explore modifications to these models with additional input information and their use in production environments. The machine learning model uses the U-Net architecture \citep{ronneberger2015u_unet} with the MobileNet-v3 \citep{howard2019mobilenetv3} network as the encoder. As inputs, we primarily use a combination of RGB bands and one of the selected matched filter products - MF, Mag1c, or WMF.
We also explore encoding the wind and location information as additional input bands. This is motivated by the practice established by the works of \citep{lang2023high, vaughan2024MARS_S2L_model}. 
The model and its input products are illustrated in Figure \ref{fig:model_illustration}.

We use the Segmentation Models Pytorch (SMP) Python library \citep{SMP_library} with default hyperparameters for these models. 
More concretely, we use an an encoder depth of 5 with the default number of channels in each decoder block (from first to last: 256, 128, 64, 32, 16). The decoder uses batch normalisation layers (in each block, after each Conv2D layer).
Importantly, our models can be reproduced exactly using our open-source code. Additionally, we provide an architecture ablation study in appendix A. materials in Section \ref{sec:ablation_architectures}.

As was initially explored in the work of \citep{STARCOP}, models trained on matched-filter products can generalise to data across sensors. Consequently, we favor them here over models trained on spectral radiance \citep{HyperspectralViTs}.  The latter can potentially exceed the matched filter's ability to reject false positives, but are trained for the spectral sampling of a specific sensor and do not easily generalize across instruments.  For operational simplicity over long deployments, we prefer to use enhancement fields, which provide the ability to handle new instruments coming online and robustness to minor changes in input radiance caused by calibration drift or data processing versions. In this unique situation of using three datasets from three distinct hyperspectral sensors, we explore zero-shot generalisation and fine-tuning techniques. Specifically, we explore whether models trained on the largest dataset of events in EMIT can aid in the detection of methane leaks, also in data from the other two sensors.

We note that the sampling strategy for EMIT, EnMAP, and PRISMA data differs (temporal for EMIT, and spatial split for the latter two). We consider temporal splits to be more faithful to the operational use case (a dataset of prior emissions used to train models later deployed on more recent data); however, this was not possible for the smaller EnMAP and PRISMA datasets. 
As long as these splits contain diverse samples, different sampling strategies won't lead to significantly different results; as such, we don't expect to see problems with model generalisation. Furthermore, events observed by different sensors are necessarily going to differ, as the capture of these events by different satellite missions occurs at different time points.

Finally, we explore the robustness of the proposed machine learning models for their deployment in production pipelines. For this, we extend the model evaluation to full granules rather than restricting it to the small chips in the test set.
These constitute what is referred to as in-the-wild data in Computer Vision. In-the-wild data typically have a different distribution and class balance than the training set, with many more negative cases. This can lead to a very large number of false-positive detections. To reduce the number of false-positive detections in full tiles, we use model ensembles. For reporting results, it is customary to train the same machine learning model with several repetitions, each with differently initialised weights, and this is used to report the average and standard deviation of the scores. Here, we can, however, treat these repeatedly trained models as an ensemble. Model ensembles have been used to improve predictive capabilities and in the uncertainty estimation literature, such as \citep{beluch2018power, ruuvzivcka2020deep}. Intuitively, different models in the ensemble will have learned distinct internal representations of the data, all compatible with a limited training dataset. When presented with a new, potentially out-of-distribution sample, they will provide different predictions on uncertain data. Using the ensemble's average output has a regularizing effect, reducing the impact of initial weights on the prediction. The suggestion to use model ensembles for improved methane leak detection has been identified as a potential future research direction in \citep{ruuvzivcka2025intelligent}.

\subsection{Matched filter thresholding baselines}

As baseline methods, we use thresholding of the various methane enhancement products, followed by morphological operations, to reduce the so-called salt-and-pepper noise in the predictions. The same practice has been explored in \citep{STARCOP} and in \citep{HyperspectralViTs}. In the latter, different kernel sizes were used for the morphological operations. We tried all variants and report only the best performing ones.
Specifically, we threshold the matched filter products by 500 ppm$\times$m, and follow with erosion and dilation morphological operations with one of two 3x3 kernels - ``ones'' (all values set to ones) or ``cross'' (middle row and middle column full of ones).

\section{Experiments}\label{sec:experiments}

In this paper, we explore two main types of experiments. 
First, we explore a scenario in which we have a large dataset of methane leak events and want to select the best-performing model variant. Broadly speaking, this effort will be a follow-up to the work presented in \citep{STARCOP}; however, we add several innovations to the models in terms of the methane enhancement products used, input data modalities, and, finally, methods useful for final deployment, such as model ensembling. 
Secondly, we explore the generalisation ability of the models trained in the first step, with the aim of adapting the machine learning models to data from different hyperspectral sensors. Methodologically, we adapt the steps taken in \citep{HyperspectralViTs}, where models were pre-trained on a large dataset (in that case, synthetic data), to be later fine-tuned and evaluated on much smaller real-world datasets. In our scenario, we adapt models pre-trained on the larger EMIT dataset to the smaller datasets of EnMAP and PRISMA
sensors.

As was also the case in previous works \citep{STARCOP}, the data is heavily imbalanced (there are more pixels corresponding to the background class than to the in-plume class). We address this with the following two techniques. Firstly, when loading data batches, we use the PyTorch ``WeightedRandomSampler'' to oversample tiles from the minority class. Secondly, for each tile, we weight the loss (before computing the mean over each pixel and tile in the batch) using adjusted matched-filter products. We take the product used by the model (e.g., WMF) and clip it to the range 0.1 and 1.0. Doing so, the pixels that correspond to high values in the matched filter product (either true plume pixels or strong confounder background pixels) have higher weights than the rest of the scene; as such, the model is guided to learn to differentiate between them. Also note that no-data pixels are masked out in the losses.
Our models use binary cross-entropy loss with a sigmoid layer as the final activation function.

\subsection{Training on EMIT datasets}

Our training hyperparameters are informed by the work of \citep{STARCOP}. Our models use a similar U-Net architecture, in our case with the MobileNet-v3 encoder. As inputs, we use the RGB channels of the hyperspectral data, converted to reflectances, alongside one of the computed methane enhancement products. In addition, we optionally provide the model with 2 additional bands of data capturing wind information (wind direction vectors at the center of each tile) and 3 additional bands of data encoding location information (latitude and longitude values, represented following the approach of \citep{lang2023high}). 
In our experiments, we compare which methane enhancement products are best for our trained models, and whether the additional information from wind and location helps model training.

We train these models on the EMIT dataset, which has three subsets: train, validation, and test. The validation dataset is used to select the best-performing checkpoint. In all experiments, we train the models for 60 epochs using the Adam optimizer with an initial learning rate of 0.001. From our training dataset, we extract 128x128 px tiles with a 64 px overlap. We augment these with random rotations, spatial jitter of 16 px, and random vertical and horizontal flips to prevent methane leak events from appearing in the same location during training. We use a batch size of 64 tiles. When evaluating, we process tiles in their original resolution - this is 256x256 px for the regular dataset and variable resolution of about 1280x1242 px in the case of the full tile data (when needed, we pad this data with zeros to a multiple of 32).

Finally, in our evaluation, we also explore model ensembles. In all experiments, we repeat the training process 5 times to obtain the average and standard deviation for each output. For model deployment, we make predictions with all models and then average them as the final ensemble output. 
Each model in the ensemble uses the same architecture but is trained from a different random initialization and undergoes a stochastic training run.
We also explored approaches to weight averaging or methods to assess agreement between the models, but obtained worse results. While model ensembles increase the computation time for model inference (linearly with the number of models used), the model prediction is very fast compared to other steps in the workflow, such as downloading data from NASA servers. 
We have timed individual steps in the pipeline on a consumer grade computer (MacBook Pro, Nov 2024, Apple M4 Max with 64 GB RAM) - 1.) model loading (occurs once) takes 0.9s, 2.) downloading data from the NASA's Earthdata system (on a 260Mbps connection) takes about 130s, 3.) data loading, orthorectification and computation of WMF takes about 20s, 4.) and finally model inference takes only about 1.6s (this step is approximately 5x slower when using ensemble of 5 models).
As such, model inference does not create a computational bottleneck in the on-ground prediction scenario (as opposed to model predictions performed directly on satellites, as explored in \citep{HyperspectralViTs}).

\subsection{Adaptation to PRISMA and EnMAP data}

In the next set of experiments, we use the best models obtained from EMIT training and explore how to adapt them for data from a different hyperspectral sensor. Our models use the same input modalities and due to data conversion into reflectances, the data is in similar value ranges. This makes data normalisation simpler. We also note that we can compute the same methane enhancement products from a variety of hyperspectral sensors - for that reason, we are able to use the same computed WMF product (selected as the best performing MF variant).

Our datasets from the PRISMA and EnMAP sensors are smaller and less diverse than the dataset of EMIT. For this reason, we compare (1) using models pre-trained on EMIT data in a zero-shot learning manner, with (2) training new models from scratch on the smaller datasets, and finally (3) with model fine-tuning. When fine-tuning, we can either use the data from the target sensor only (here PRISMA or EnMAP), or we can create a combined dataset from the training datasets of the EMIT and one target sensor.

When creating these smaller datasets, we chose to only split data in between train and test subsets. This enables us to use more of the data we have for training, but prevents us from selecting the best model with a validation dataset. However, we can still select the best performing models and treat the data which these models will be used with in the future as the true test datasets. Otherwise, we can always use the last checkpoint after a selected number of epochs.

Models trained from scratch use the default learning rate of 0.001 consistent with the hyperparameter settings used when training on the EMIT dataset - and are trained for 50 epochs. On the other hand, for fine-tuning models, we select a much lower learning rate of 
$10^{-6}$ and fine-tune them for 10 epochs, while evaluating at the end of each one to detect if the models are overfitting.

\section{Results}\label{sec:results}

In this paper, we present four main results in the following structure. Table \ref{tab:emit_results} and section \ref{sec:res_input_products} explores different input modalities and their influence on the model performance. Section \ref{sec:res_fulltiles} further explores the evaluation of these models on full scene data and details our proposed improvements of using model ensembles to reduce false alarms. Tables \ref{tab:enmap_results} and \ref{tab:prisma_results} and section \ref{sec:res_adaptation} covers adaptation between sensors. Finally, section \ref{sec:res_deployment} and \ref{sec:res_mititgation} details the deployment of our models in the real production pipelines used at UNEP's IMEO.

\subsection{Methane detection in EMIT data}
\label{sec:res_input_products}

Table \ref{tab:emit_results} shows results of our baselines and proposed models comparing different input modalities, including three types of matched filter product and auxiliary data such as the wind and location information. When considering all baseline approaches, we see some variation between the different input modalities. However, all variants perform worse than any of the trained models. This is consistent with prior research and highlights the benefits that can be gained by using deep learning models for this task.

\begin{figure}[!h]
    \centering
    \includegraphics[trim={0 0 0 0.8cm},clip,width=1.0\linewidth]{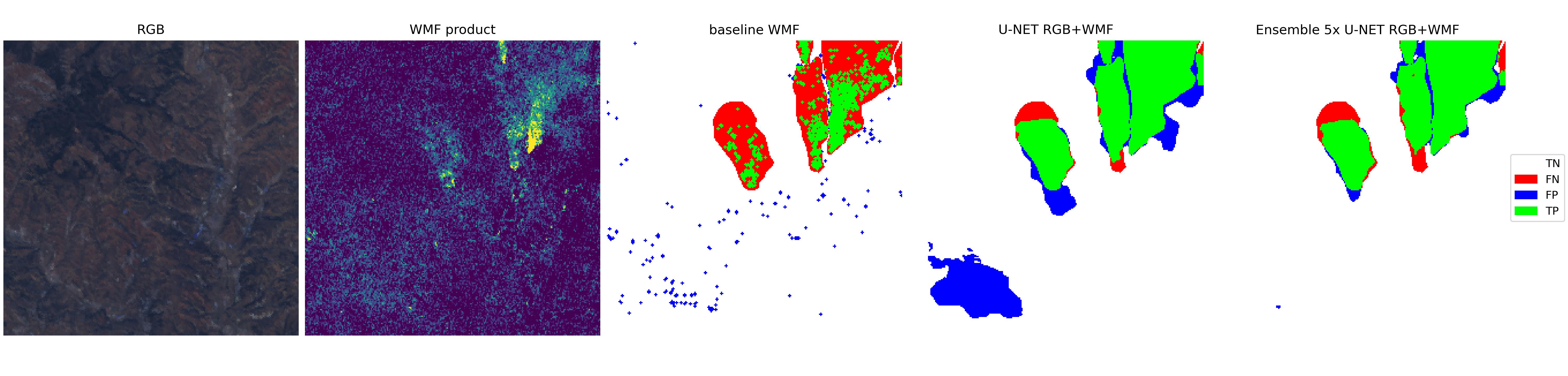}
    \includegraphics[trim={0 0 0 0.8cm},clip,width=1.0\linewidth]{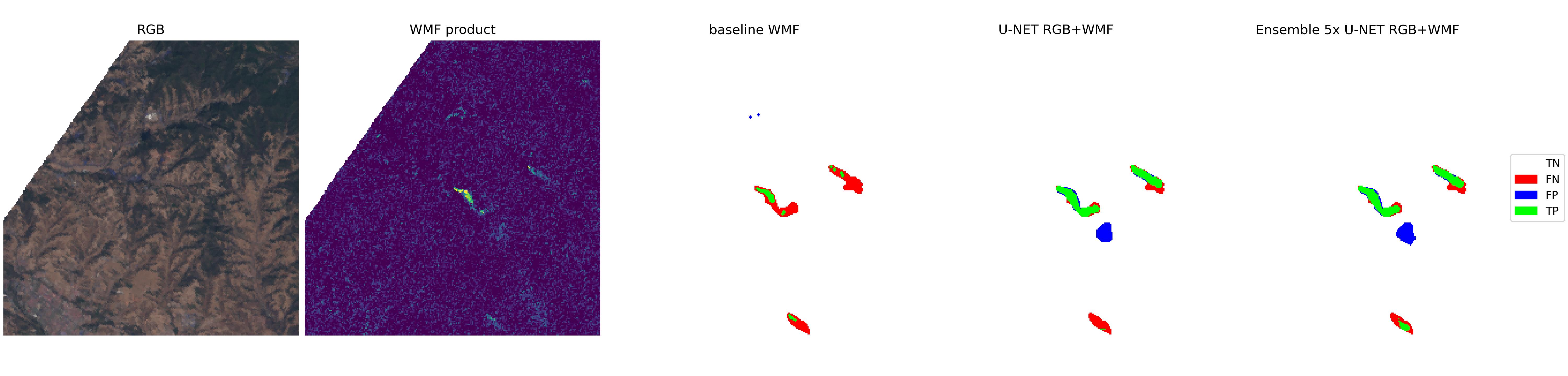}
    \includegraphics[trim={0 0 0 0.8cm},clip,width=1.0\linewidth]{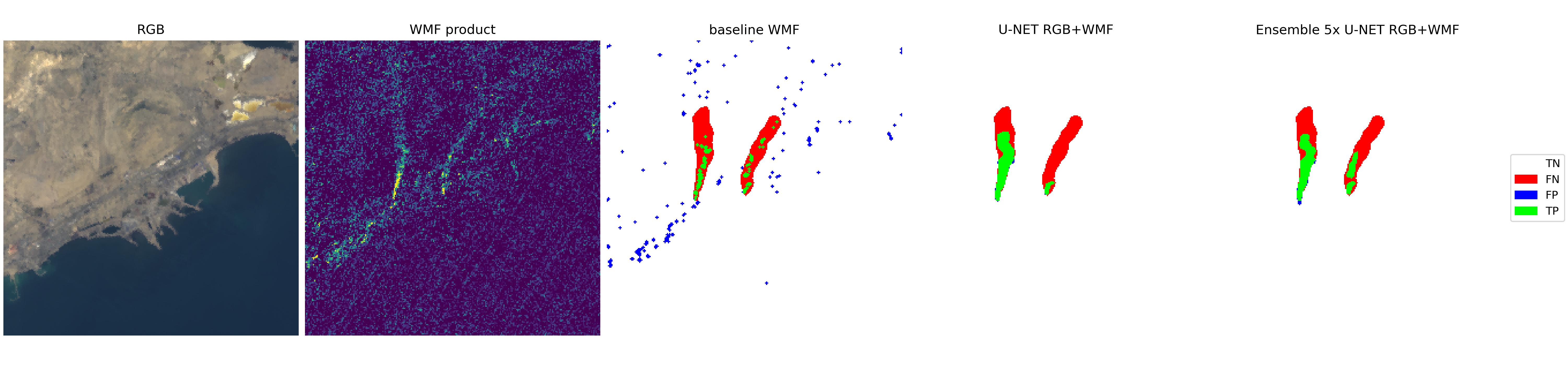}
    \caption{Qualitative results comparing our machine learning models on samples from the EMIT test dataset. Shown from left to right are predictions from the baseline (using WMF, 500 ppm$\times$m as threshold and the ``cross'' kernel), single U-Net model (using RGB+WMF) and finally an ensemble of five of these models.}
    \label{fig:EMIT_smalltiles}
\vspace{-2mm}
\end{figure}

When comparing models with RGB and single methane enhancement product – the default MF, Mag1c or WMF – we observe interesting improving relation. If we focus just on the average F1 score in Table \ref{tab:emit_results}, the Mag1c variant performs the worst with 43.59, default MF gets 53.92, while the WMF is the best performing product with 63.07. Between the worst and the best, this is a significant score increase by over 44\%. It seems that the WMF, with its improved background rejection from using a broader spectral interval, provides better target/background contrast that simplifies the learning task.  Figure \ref{fig:EMIT_smalltiles} shows the qualitative comparison between the baseline method and our proposed models.

The inclusion of auxiliary data, such as the wind and location information, further boosts the performance of our best models, however these gains are only minor in comparison with the selection of the best matched filter product. Wind information encoding boosts the average F1 score to 63.68 (improvement by just 0.61 points) and location encoding boosts this to 64.40 (improvement of additional 0.72 points). Other scores are also improved in similar magnitude.

\subsection{Evaluation on full EMIT tiles}
\label{sec:res_fulltiles}

\begin{figure*}[!h]
    \centering
    \includegraphics[width=1.0\linewidth]{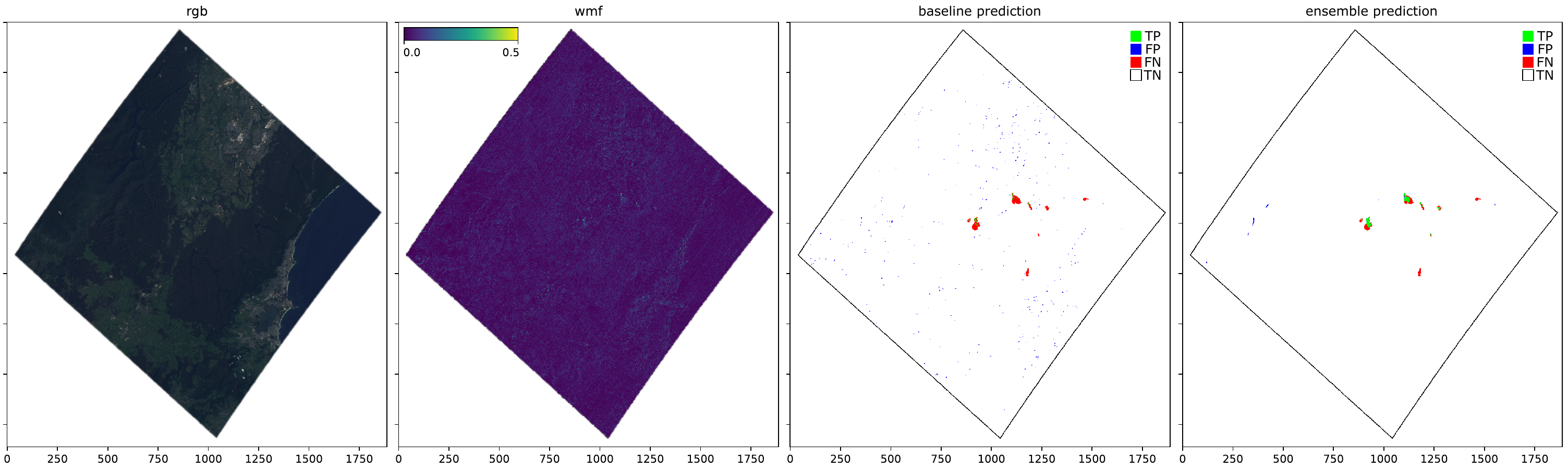}
    \caption{EMIT full tile example prediction comparing the thresholding baseline (one using WMF and the ``ones'' kernel) with the ``Ensemble, 5x U-Net RGB+WMF\fixS{, Wind}'' model. Note the so-called salt-and-pepper noise of the baseline, this is present even after the morphological operations. While we note that it could be further reduced by using larger kernels, true detection of real plume events would also suffer. 
    WMF values are scaled as inputs to the model (0.5 corresponds to 4000 ppm$\times$m). For better legibility, we recommend viewing the web version of this article.}
    \label{fig:emit_full_tile_preds}
\vspace{-2mm}
\end{figure*}

Table \ref{tab:emit_results} further details evaluation of our models on the full tile test sets with the number of detected and missed methane leak events and the number of false alarms. Note that for EMIT data, we created the full tile dataset from 105 source granules with known plume events and 95 granules with no events. An example full tile evaluation is shown on Figure \ref{fig:emit_full_tile_preds}. As a successful detection in fulltile evaluation, we count any prediction that overlaps with a ground truth label in the data. When multiple predictions overlap with one larger ground truth label, we consider this as only one detection (this can occur in cases when the plume shape is complex and for example the plume origin and plume tail get predicted as two separate shapes).

Switching from any baseline method to deep learning produces significant improvements. While the WMF baseline has over 79k false alarms, this number is reduced to just over 5.6k (using the U-Net RGB+WMF, Wind variant).

In this context we also evaluate our proposed approach of using model ensembles. This further boosts the tiled dataset statistics, achieving the best performing models in terms of scores like F1 and AUPRC. More importantly, it dramatically reduces the number of false alarms created by these models, while achieving one of the largest number of detected events (using the variant with the wind information).

We highlight that while the improvements for example over the F1 score remain relatively minor when using ensembles (by 2.84\%) the main contribution is in the reduction in the number of false alerts by more than 97\% over the classical baselines. This is particularly relevant for model deployment in a real production pipeline, an application which has a low tolerance for false positive errors.

\begin{table*}[]
\caption{Results on EMIT. We show the average of training 5 runs of our models.} \label{tab:emit_results}
\centering
\scalebox{0.88}{
\begin{tabular}{@{}llllllll@{}}
\toprule
\textbf{EMIT}                                 & \multicolumn{4}{l}{Tiled dataset, segmentation:}                        & \multicolumn{3}{l}{Fulltile sources, individual events:} \\
Model variant & AUPRC & F1 & Precision & Recall & Detected & Missed & False Alarms \\ \midrule
baseline Mag1c (thr. 500, cross) & N/A & 22.45 & 36.22 & 16.27 & 187 & 124 & 41903 \\
baseline MF (thr. 500, ones) & N/A & 27.7 & 39.61 & 21.29 & 217 & 94 & 110383 \\
baseline WMF (thr. 500, cross) & N/A & 23.4 & 38.8 & 16.75 & \textbf{225} & \textbf{86} & 79273 \\ \midrule
U-Net RGB+Mag1c \citep{STARCOP} & 40.87 ±2.75 & 43.59 ±3.17 & 53.48 ±1.90 & 37.00 ±4.36 & 137 ±21 & 174 ±21 & 5789 ±2157 \\
U-Net RGB+MF & 56.06 ±2.49 & 53.92 ±2.14 & 63.04 ±3.98 & 47.51 ±4.31 & 203 ±20 & 108 ±20 & 6677 ±2138 \\
U-Net RGB+WMF & 68.26 ±3.06 & 63.07 ±2.46 & 72.99 ±1.17 & 55.65 ±3.82 & 213 ±20 & 98 ±20 & 3565 ±1830 \\
U-Net RGB+WMF, Wind & 68.97 ±3.16 & 63.68 ±2.77 & 69.28 ±2.75 & \textbf{59.01 ±3.69} & 220 ±19 & 91 ±19 & 5644 ±2570 \\
U-Net RGB+WMF, Wind, Location & 69.44 ±2.60 & 64.40 ±2.08 & 71.88 ±2.35 & 58.65 ±4.66 & 191 ±22 & 120 ±21 & 2403 ±899 \\ \midrule
Ensemble, 5x U-Net RGB+WMF & \textbf{73.58} & 65.13 & \textbf{79.87} & 54.98 & 207 & 104 & \textbf{1501} \\
Ensemble, 5x U-Net RGB+WMF, Wind & 72.64 & \textbf{65.49} & 73.81 & 58.85 & 216 & 95 & 2118 \\ \bottomrule
\end{tabular}
}
\end{table*}

\subsection{Adaptation to other sensors, EnMAP and PRISMA}
\label{sec:res_adaptation}

Tables \ref{tab:enmap_results} and \ref{tab:prisma_results} present our results when evaluating model adaptation on data from  EnMAP and PRISMA. In both cases, the number of methane leak events is much smaller, and as has been explored in prior research \citep{STARCOP, HyperspectralViTs, mancoridis2025multi}, zero-shot generalisation and model fine-tunning can be used to transfer knowledge learned on the larger source dataset (in our case composed of data from EMIT).

Generally, we observe relatively high zero-shot performance of the out-of-the-box models pre-trained on the EMIT data used for both EnMAP and PRISMA cases. These seem to outperform simple training from scratch approaches, which might be due to the relatively small number of methane leak events present in these additional datasets. Fine-tuning of these models usually boosts the performance further, but depending on which sensor is used, we observe different behaviour when using just the target domain, or a combination of the two datasets (source and target).

\begin{table*}[]
\caption{Results on EnMAP. We show the average of training 5 runs of our models.} \label{tab:enmap_results}
\centering
\scalebox{0.88}{
\begin{tabular}{@{}llllllll@{}}
\toprule
\textbf{EnMAP}                           & \multicolumn{4}{l}{Tiled dataset, segmentation:}                              & \multicolumn{3}{l}{Fulltile sources, individual events:} \\
Model variant & AUPRC & F1 & Precision & Recall & Detected & Missed & False Alarms \\ \midrule
Baseline (WMF 500 thr, ``ones'' kernel) & N/A & 35.37 & 37.82 & 33.21 & \textbf{116} & \textbf{21} & 20264 \\ \midrule
Train from scratch, 50 ep. & 43.24 ±3.45 & 44.39 ±2.62 & 56.43 ±6.91 & 36.85 ±1.65 & 70 ±6 & 67 ±6 & 2132 ±843 \\ \midrule
Zero-shot (EMIT U-Net RGB+WMF model) & 62.46 ±1.83 & 56.84 ±1.57 & 50.61 ±3.04 & 65.16 ±3.32 & 103 ±8 & 34 ±8 & 2514 ±1522 \\ \midrule
Fine-tune (EnMAP), 1 ep. & 59.87 ±5.84 & 55.32 ±4.36 & \textbf{68.48 ±4.5} & 46.69 ±5.47 & 93 ±7 & 44 ±7 & \textbf{1013 ±342} \\
Fine-tune (EMIT+EnMAP), 4 ep. & 64.49 ±1.6 & 59.28 ±1.55 & 55.17 ±2.98 & 64.23 ±1.97 & 103 ±5 & 34 ±5 & 1951 ±811 \\ \midrule
Ensembled Zero-shot & 68.71 & 61.65 & 58.5 & \textbf{65.15} & 102 & 35 & 1302 \\
Ensembled Fine-tune (EMIT+EnMAP), 4 ep. & \textbf{69.94} & \textbf{64.02} & 63.51 & 64.54 & \fix{104} & 33 & 1115 \\ \bottomrule
\end{tabular}
}
\end{table*}

More concretely, in Table \ref{tab:enmap_results} we see progressive improvement in the F1 score when comparing training from scratch on EnMAP data (44.39 F1), zero-shot deployment of the EMIT model (56.84 F1) and finally fine-tuning of EMIT models on just the EnMAP data (55.32 F1), or on an extended dataset of both EMIT and EnMAP (59.28 F1).
Ensembling of the best models boosts this performance to final F1 score of 64.02.
In total we gain over 44\% when using the ensembled fine-tuned model over model depending only on the EnMAP data, or by 12.6\% when comparing against only zero-generalised EMIT model.

\begin{figure}[!h]
    \centering
    \includegraphics[trim={0 1.5cm 0 0.8cm},clip,width=1.0\linewidth]{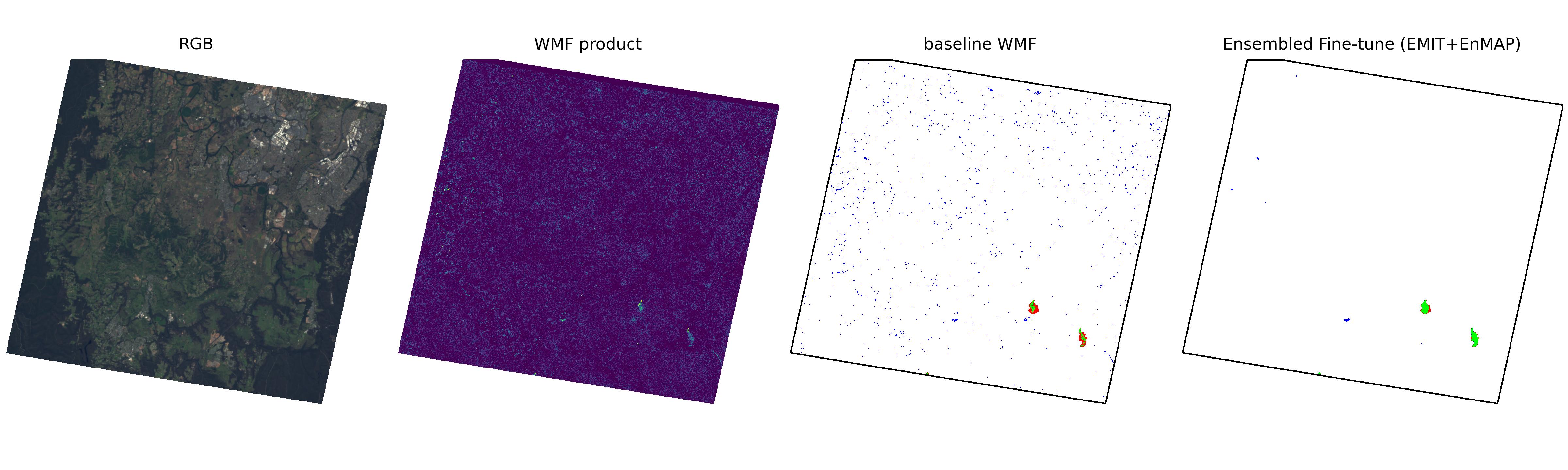}
    \includegraphics[trim={0 1.5cm 0 0.8cm},clip,width=1.0\linewidth]{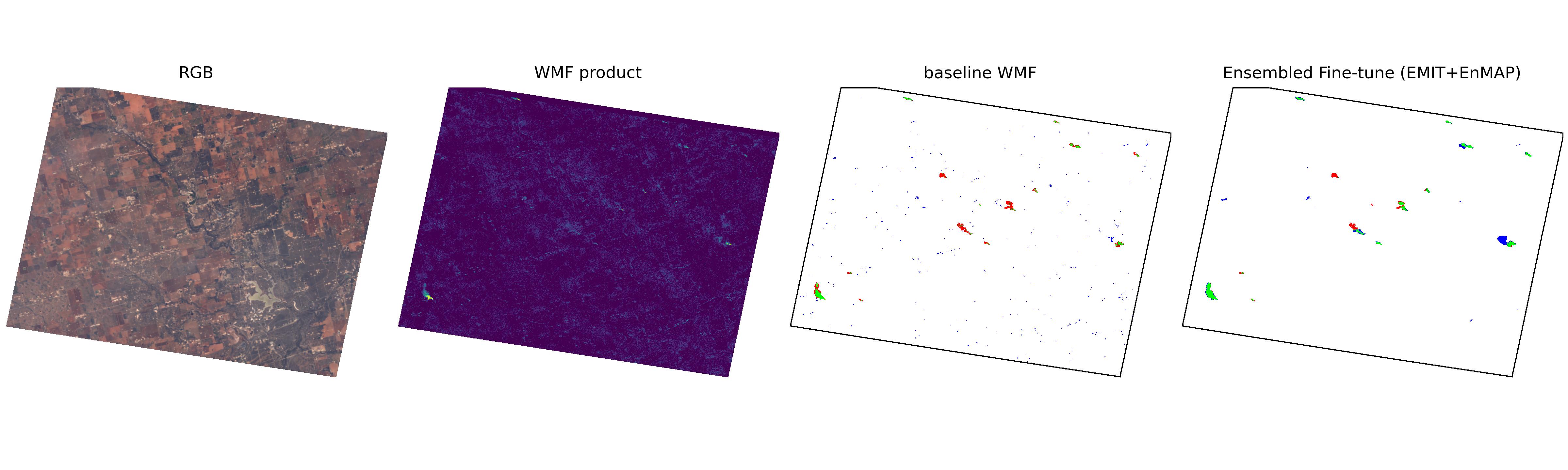}
    \caption{Model prediction examples on the EnMAP data comparing baseline approaches with our proposed Ensemble models (showing the ``Ensembled Fine-tune (EMIT+EnMAP)'' model).}
    \label{fig:EnMAP_fulltiles}
\vspace{-2mm}
\end{figure}

Here we note the importance of evaluating models on datasets with full tile samples. There, the number of detected events varies depending on the used generalisation method.
Simple zero-shot generalisation of the EMIT model outperforms variants that would fine-tune only on the much smaller dataset of EnMAP images.
If we instead fine-tune on a dataset constructed by combining the EMIT and EnMAP training datasets, we receive the best from the both approaches - previously discussed improvements in the F1 score and also maintained large number of detected events.
We also observe reduced number of false positives.
For best performance, we can again leverage ensemble models. These reduce the number of false positives by more than 55\% while maintaining the same number of detected events (comparing zero-shot against our best ensemble model).
Qualitative results can be seen on example scenes in Figure \ref{fig:EnMAP_fulltiles}.

\begin{table*}[]
\caption{Results on PRISMA. We show the average of training 5 runs of our models. } \label{tab:prisma_results}
\centering
\scalebox{0.88}{
\begin{tabular}{@{}llllllll@{}}
\toprule
\textbf{PRISMA}                           & \multicolumn{4}{l}{Tiled dataset, segmentation:}                              & \multicolumn{3}{l}{Fulltile sources, individual events:} \\
Model variant & AUPRC & F1 & Precision & Recall & Detected & Missed & False Alarms \\ \midrule
Baseline (WMF 500 thr, ``ones'' kernel) & N/A & 26.05 & 23.14 & 29.80 & \textbf{123} & \textbf{20} & 57270 \\ \midrule
Train from scratch, 50 ep. & 25.91 ±2.52 & 31.55 ±2.56 & 31.0 ±5.32 & 33.61 ±4.94 & 90 ±16 & 53 ±16 & 5032 ±1264 \\ \midrule
Zero-shot (EMIT U-Net RGB+WMF model) & 31.84 ±1.99 & 40.5 ±2.9 & 31.57 ±3.22 & \textbf{57.18 ±4.89} & 111 ±13 & 32 ±13 & 4984 ±2727 \\ \midrule
Fine-tune (PRISMA), 6 ep. & 37.05 ±4.86 & \textbf{44.37 ±2.09} & 40.32 ±4.73 & 50.14 ±2.97 & 101 ±7 & 42 ±7 & 2806 ±1072 \\
Fine-tune (EMIT+PRISMA), 2 ep. & 33.0 ±1.25 & 42.11 ±1.16 & 34.36 ±1.77 & 54.71 ±3.35 & 107 ±8 & 36 ±8 & 3715 ±1665 \\ \midrule
Ensembled Zero-shot & 35.6 & 44.04 & 37.02 & 54.36 & 113 & 30 & 2881 \\
Ensembled Fine-tune (PRISMA), 6 ep. & \textbf{40.61} & 43.61 & \textbf{43.11} & 44.13 & 103 & 40 & \textbf{1674} \\
Ensembled Fine-tune (EMIT+PRISMA), 2 ep. & 36.08 & 43.79 & 37.97 & 51.72 & 106 & 37 & 2248 \\ \bottomrule
\end{tabular}
}
\end{table*}

\begin{figure}[!h]
    \centering
    \includegraphics[trim={0 0.5cm 0 0.2cm},clip,width=1.0\linewidth]{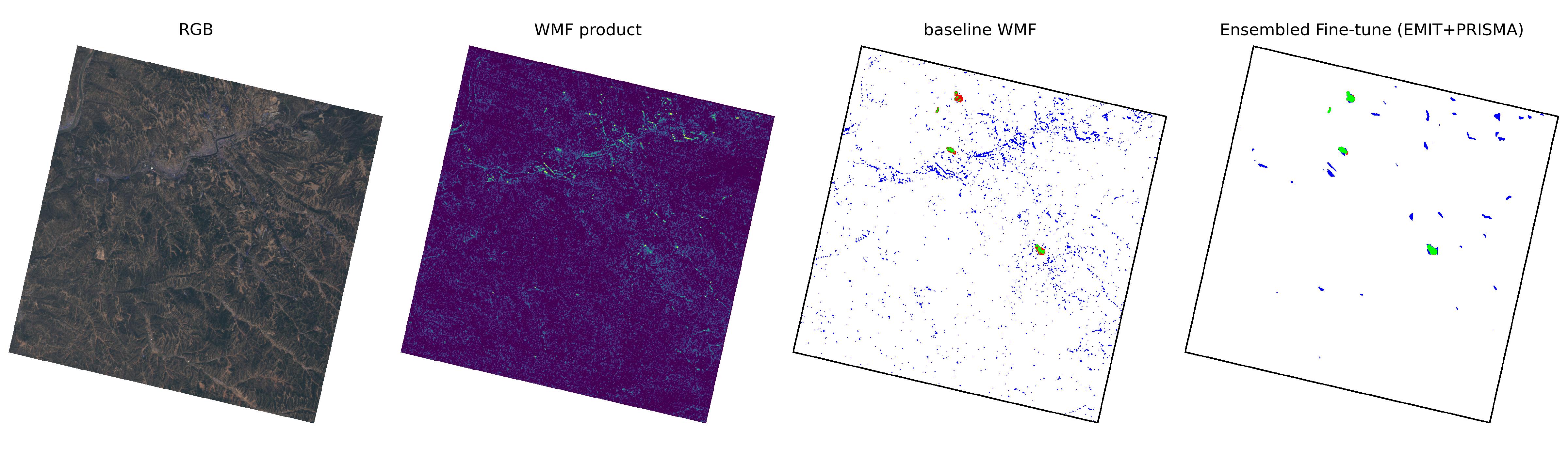}
    \caption{Model prediction examples on the PRISMA data comparing baseline approaches with our proposed Ensemble models (showing the ``Ensembled Fine-tune (EMIT+PRISMA)'' model).}
    \label{fig:PRISMA_fulltiles}
\vspace{-2mm}
\end{figure}

As with the EnMAP dataset, when adapting our models for the PRISMA data in Table \ref{tab:prisma_results}, we observe improvements in the F1 score when going from training from scratch (31.55 F1) to zero-shot deployment of the EMIT model (40.5).
Finally, when fine-tuning the models on the PRISMA dataset we get a 44.37 F1 score.
When fine-tuning on a dataset built from a combination of EMIT and PRISMA training datasets we get an F1 score of 42.11. 
There is a small difference with the EnMAP scenario, where models trained on the combined dataset had better performance, while we see slightly better performance when using the PRISMA only data.
This may be due to the larger size of the available PRISMA data in contrast to the small EnMAP dataset which may lead to model overfitting when used in isolation.
As before, model ensembles obtain a high F1 score (above 43), while maintaining large number of detected events and smaller number of false positive detections. Figure \ref{fig:PRISMA_fulltiles} shows an example full tile scene.
Interestingly, if we wanted to use the model that detects the most events, we would chose either the model pretrained on EMIT only, or models fine-tuned on a combined dataset of PRISMA and EMIT images.

\subsection{Results from operational deployment}
\label{sec:res_deployment}

\begin{table}[h]
\caption{Operational deployment results. The numbers correspond to the statistics after \fix{\howmanymonthsdeployedN}months of model deployment. As this is a living system, these numbers increase daily.}
\label{tab:op_dep}
\centering
\scalebox{1.0}{
\begin{tabular}{@{}lccc@{}}
\toprule
                                  & \textbf{EMIT} & \textbf{EnMAP} & \textbf{PRISMA} \\ \midrule
\textbf{Processed tiles}          & 15,968  &  7,478  & 1,578    \\
\textbf{Verified detected plumes} & 1,990  &    659   &    202       \\
\textbf{Different countries}      & 68    &   46   &     27      \\
\textbf{Plumes notified}          &  642   &   120   &   72       \\ 
\bottomrule
\end{tabular}
}
\end{table}

Table \ref{tab:op_dep} presents the results of using our proposed models in the operational detection pipeline used at the Methane Alert and Response System (MARS) at the United Nations Environmental Program (UNEP).

\fix{Additional analysis of the operational deployment of our system is also detailed in appendix B. in Section \ref{sub_operational}.}

Internally within the UNEP's IMEO, a tool called Plume Viewer is used to collect information from multiple sensors over areas informed by previously known emission locations and vector maps of relevant assets. There, a team of analysts searches through daily monitoring data and manually delineates and verifies the methane leak events. Needless to say, this process is quite labour intensive and can tremendously benefit from even semi-automated detection tools. For that purpose, our models were developed to perform semantic segmentation as a pre-processing step that aids with this manual search process.

As for the models we use in deployment, we chose the ensemble of 5 U-Net RGB+WMF models and use it for all sensors given their high generalisation capabilities demonstrated in Tables \ref{tab:enmap_results} and \ref{tab:prisma_results}. We do not use models informed by location despite the small improvements in number of false alerts observed in  Table \ref{tab:emit_results}, as this information may lead to the model remembering locations of prior emissions and to poor generalisation capabilities to new, previously unobserved areas (as these would appear as out-of-distribution data for the models). Nonetheless, analysts have access to both the location (with the ability to see scenes of previous overpasses) and the wind vector.

\begin{figure*}[!h]
    \centering
    \includegraphics[width=1.0\linewidth]{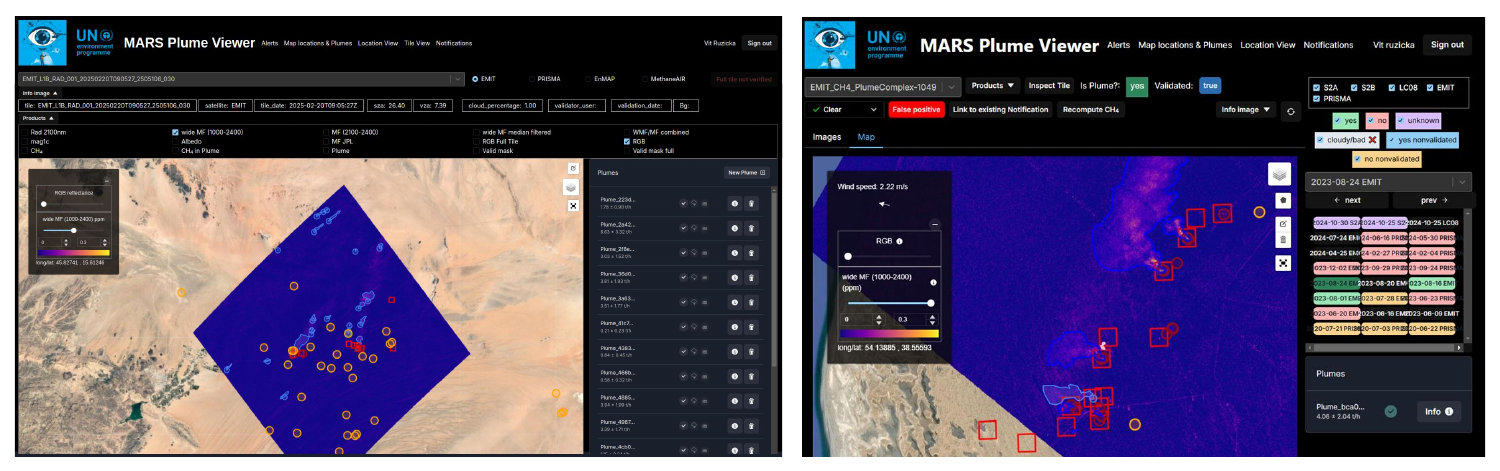}
    \caption{Example of the PlumeViewer interface used to search for methane leak events. Our models were integrated and the vectorized model prediction (in light blue) is shown for two example EMIT images. The scene on the left has not yet been manually cleaned from the false alarms, this is to demonstrate the usefulness of the system even in scenarios with false positive detections. The red squares are made around monitoring locations - these are drawn around verified plume events to help inform detections of future satellite captures.}
    \label{fig:plumeviewer}
\vspace{-2mm}
\end{figure*}

Figure \ref{fig:plumeviewer} shows the interface and examples of methane leak events proposed by our models to the human reviewers. Despite producing false positive detections, the models remain useful in guiding the reviewers towards potential methane leak events in the scene. During their review, the analyst marks the correctly detected plumes as validated (either using the model suggested boundary, or by drawing a new one as a vector) and then with one click deletes all non-validated detections. The validated detections then undergo an internal review process and are eventually notified to the relevant operators and governmental actors. 

This operational workflow is similar to the one used in \citep{vaughan2024ai}, but rather than just predicting on selected a priori known monitoring locations, we run our models on the full scenes captured by the satellite. This allows us to discover new regions of interest and to expand the map of tracked monitoring sites. For further automation, the information about the proximity of known emitters can also be used, for example by using different thresholds for model prediction in these areas. Using our models for full scene predictions is possible thanks to the insights from section \ref{sec:res_fulltiles}. 

\begin{figure}[!h]
    \centering
    \includegraphics[trim={2cm 0 2cm 0},clip,width=1.0\linewidth]{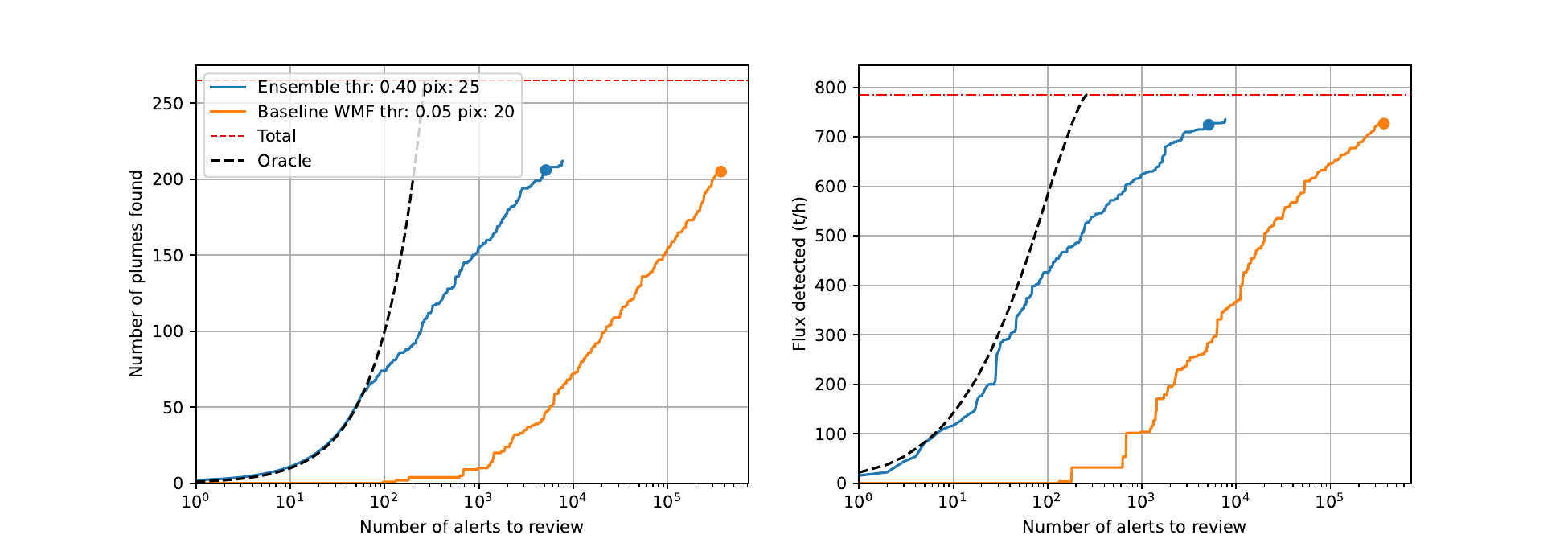}
    \caption{Deployed sorting performance using our models on events between \fix{March 2025 and February 2026} (from EMIT, PRISMA and EnMAP). Note that these are not included in our created test sets, they correspond to the new data processed throughout the operational use of our model. We show two different approaches to sorting our model predictions and highlight how many events are detected after reviewing an increasing proportion of all predicted plumes (alerts). Please note that this analysis considers only plumes from fully validated scenes, as such the number does not match the number of verified detected plumes in Table \ref{tab:op_dep} (an analyst can mark a single model prediction as verified without validating the full scene).}
    \label{fig:plume_sorting}
\vspace{-2mm}
\end{figure}

Figure \ref{fig:plume_sorting} demonstrates performance of our models for plumes between \fix{March 2025 and February 2026} \fix{focusing on the subset of fully validated scenes}. 
We additionally note that the total number of detected events differs from the full deployment results shown in Table \ref{tab:op_dep} - this is due to the fact that only fully validated scenes were used (analysts can mark a single predicted event as validated without having to validate the entire scene, which would mean deleting all the false alarm detections).
In this scenario we sort plume predictions by their assigned score. In order to compute this score we first use an initial threshold of \fix{0.4} to delineate individual plume predictions. Afterwards, we calculate the plume score as the minimum threshold of the probability map such that there are at least N=25 connected pixels inside the delineated plume (parameters for the best performing baseline are the initial threshold of 0.05 and N=20). In the figure on the left, we show on the x-axis the number of plumes that needs to be analysed in order to catch the number of plumes on the y-axis. The Oracle marks the 1:1 line (shown in log), where each analysed prediction is a true plume - this is the upper bound given a perfect knowledge. On the figure on the right we show on the y-axis the total flux of the captured plumes. We use IME method \citep{frankenberg2016airborne, varon2018quantifying_IME} to compute the flux rate. 
Model plume scores mostly consider the morphology of each plume, quantifications the estimated intensity, while scores proposed by the recent \citep{xiang2025identification} consider the spectral match with the known target signature of methane - these three likely contain complementary information.

Notably, we first deployed models for detecting methane leaks in the EMIT data, and later used them for zero shot generalisation on the EnMAP and PRISMA data\fix{ - the exact deployment timeline is available in appendix on Figure \ref{fig:deployed_over_months}}. We currently leverage the best performing model ensembles in the detection pipelines. Due to the higher cadence of observations produced by EMIT, the number of processed tiles listed in Table \ref{tab:op_dep} is higher.

\subsection{Mitigation cases}
\label{sec:res_mititgation}

\begin{figure}[!h]
    \centering
    \includegraphics[width=1.0\linewidth]{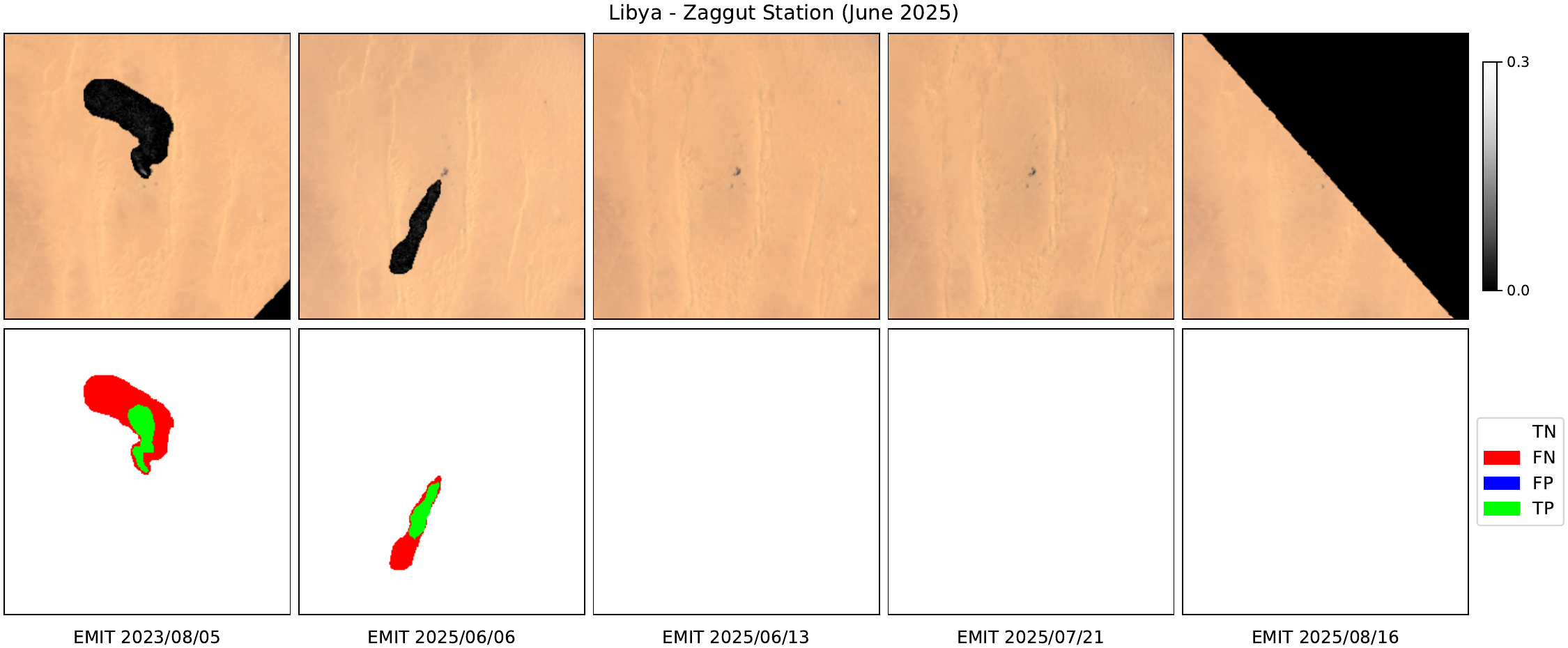}
    \includegraphics[width=1.0\linewidth]{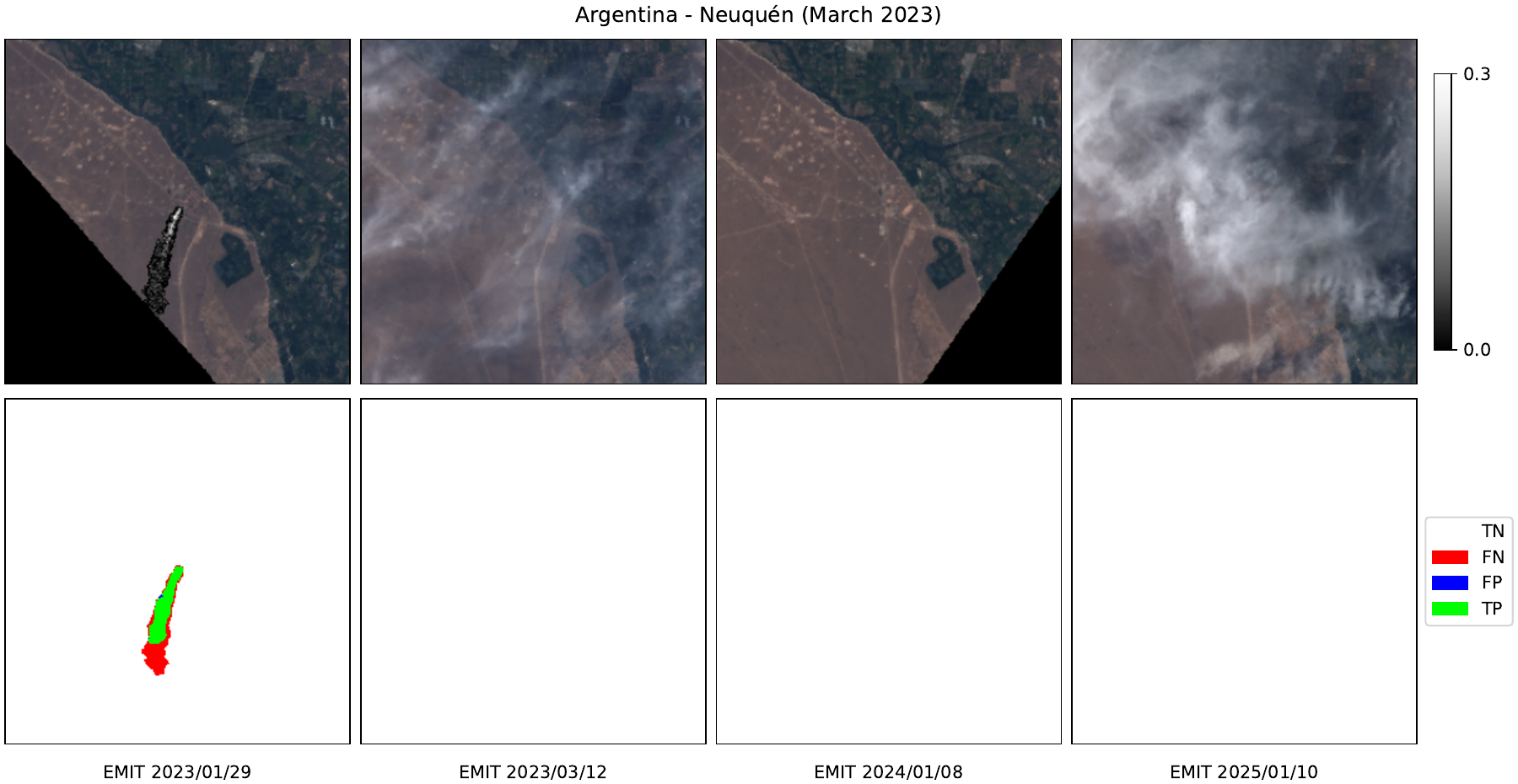}
    \includegraphics[width=1.0\linewidth]{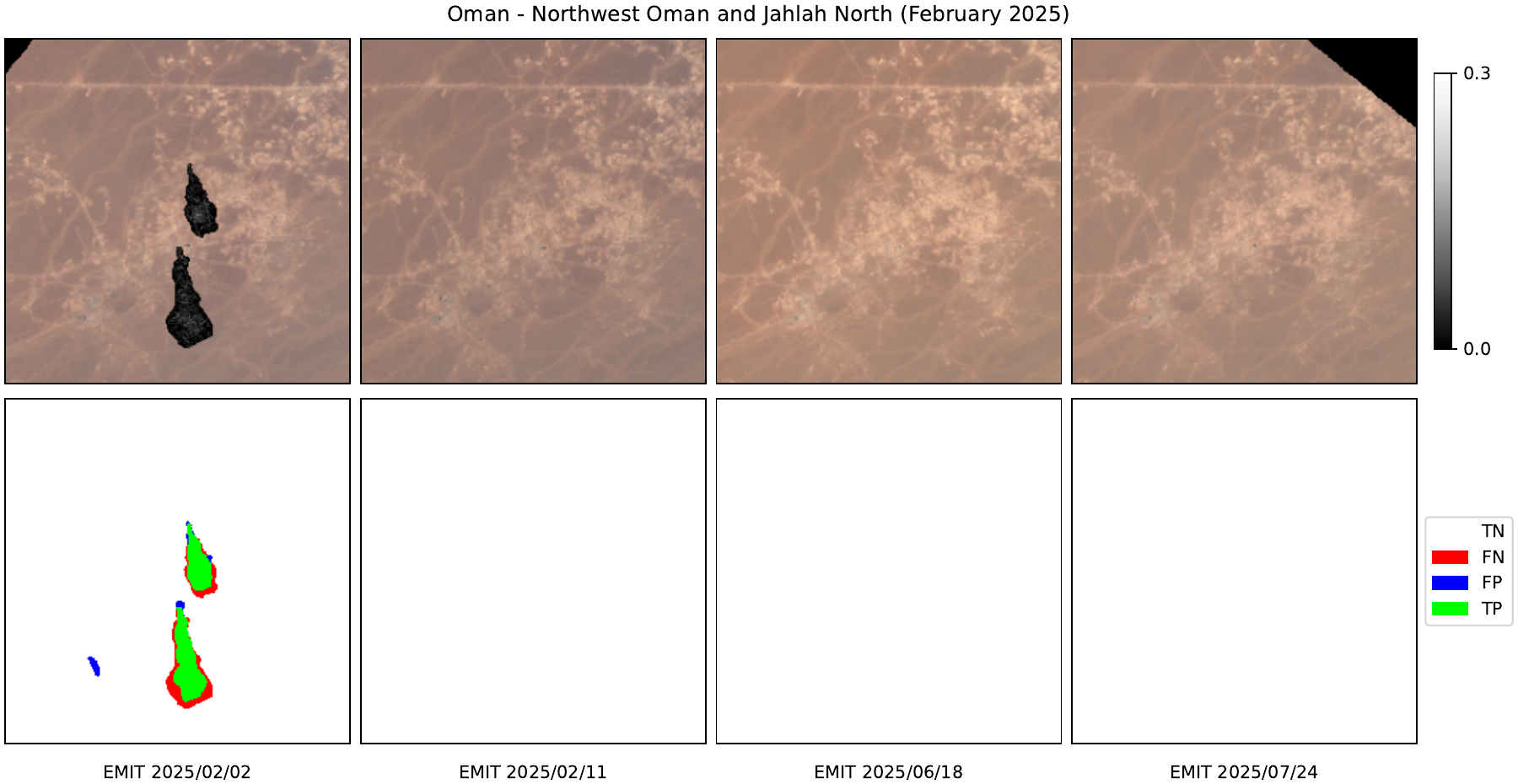}
    \includegraphics[width=1.0\linewidth]{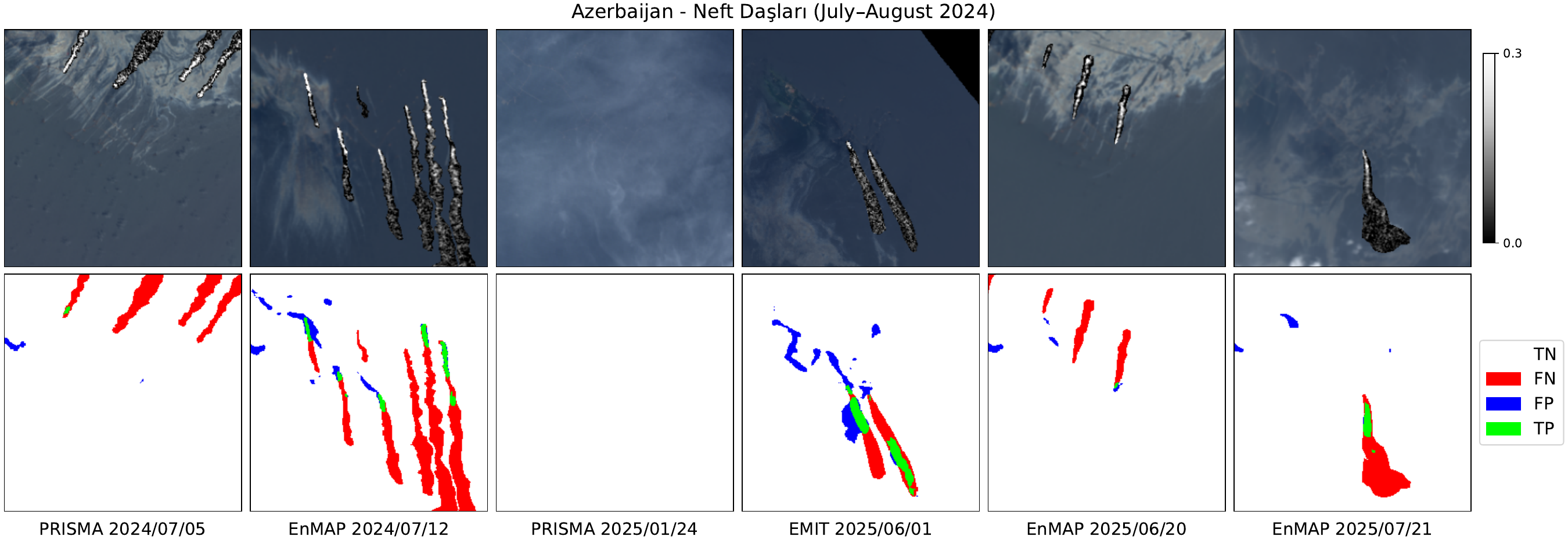}
    \caption{Mitigation case studies showing a time series of images from a selection of hyperspectral satellites with predictions of our models compared with the delineated ground truth masks. In the first row, we show an example mitigation case where our machine learning models were already operationally deployed, and aided in the notification process. In these examples, UNEP has sent MARS notifications to the relevant operators which has lead to fixing of problems that caused these leaks. The last row presents a multi-sensor example, where some parts of the overall infrastructure were fixed, while other segments were scheduled for a more concrete fix in the future. The system presented in this paper can both detect new methane emissions, but it can also be used to help with estimations of the consistency of these mitigation cases. }
    \label{fig:mitigation_cases}
\vspace{-2mm}
\end{figure}

In Figure \ref{fig:mitigation_cases}, we showcase several detections over four cases where MARS notifications triggered successful mitigation actions\footnote{These cases are listed as the MARS Mitigation Case Studies: \url{https://www.unep.org/topics/energy/methane/mars-case-studies}}. These examples demonstrate the operational use of the model to sift through the large amount of EMIT data, not only to detect potential methane plumes \fix{(using our system prospectively on new satellite scenes as shown on the first row)}, but also to confirm if mitigation actions stopped the leaks (using our system retrospectively on an archive of data as shown on the three other examples).
Figure \ref{fig:mitigation_cases}, from top to bottom, shows the detection results for mitigation cases in Libya, Argentina, Oman, and Azerbaijan before and after the undertaken actions. In Libya, using the newly deployed AI model in production, emissions at the Zaggut Station in the Waha Oil Field were detected when the wind extinguished a flare. The operator reignited it on June 6th, 2025, ending the release. In Argentina, plumes detected in March 2023 over a facility in Neuquén province prompted rapid repairs following notification. In Oman, emissions were observed on February 2nd, 2025, in the Northwest Oman and Jahlah North areas. The emissions were traced to a faulty valve and leaking flowline, both repaired within 48 hours after receiving the notification. In Azerbaijan, imagery from July–August 2024 revealed several plumes over the Neft Daşları offshore complex, after which the operator installed low-pressure gas lines to recover vented gas from two of the identified sources in a few weeks and committed to work in mitigation activities to fix the remaining notified sources in the following months. Both detections and non-detections in Fig.~\ref{fig:mitigation_cases} are consistent with feedback received by the MARS team, highlighting the value of AI-based monitoring not only for the discovery of new leaks, but also for confirming emission-free conditions and identifying potential successful mitigation interventions.

\section{Discussion and Conclusion}\label{sec:discussion}

This paper detailed the operational deployment of a machine learning system for detecting methane plumes across three imaging spectroscopy instruments: EMIT, PRISMA and EnMAP. This work was made possible by our creation of the largest publicly available dataset of manually annotated methane plumes from these sensors. We have released this dataset to the research community to serve as a benchmark and to accelerate the development of future AI models in this critical domain.

Our results demonstrate that the choice of input product can significantly impact the AI model performance.
Through a quantitative comparison, we found that models using the wide-window matched filter~\citep{roger2023wide_MF} as input achieve substantially higher F1 scores (63.07 $\pm$ 2.56) than those using either Mag1c~\citep{foote_fast_2020_mag1c} (43.59 $\pm$ 3.17) or the standard matched filter (53.92 $\pm$ 2.14) (Table~\ref{tab:emit_results}).
Furthermore, our ensemble models outperformed non-machine learning baselines by a significant margin, increasing the F1 score between 135\% to 191\% on the tiled test dataset (Tables~\ref{tab:emit_results}, \ref{tab:enmap_results} and \ref{tab:prisma_results}).

A key finding of our study is the model's strong cross-sensor generalization. Models trained on EMIT data achieved robust zero-shot performance on both PRISMA and EnMAP datasets, a capability further enhanced through fine-tuning~(Tables~\ref{tab:enmap_results} and \ref{tab:prisma_results}). This successful application of transfer learning is significant as it demonstrates that a model trained on a large, high-quality dataset from a source mission like EMIT can be rapidly deployed for new sensors with similar observational characteristics, accelerating their operational readiness.

An important lesson learned from the operational deployment is the necessity of evaluating models on full satellite granules which exposes performance issues not apparent in tiled datasets.
While our initial models produced numerous false positives on these full scenes, we found that model ensembles provided a simple yet effective approach, to reducing false positive detections by over 74\% (in comparison with previous deep learning models presented in \citep{STARCOP}).
This issue stems from the out-of-distribution (OOD) challenge inherent within our application; the training data is necessarily biased toward locations with methane-emitting infrastructure.
Consequently, when a model processes a full granule, it encounters vast areas of OOD content—such as cities, roads, rivers, and mountains—that were underepresented during training.
Ensembling mitigates this problem by smoothing over-confident predictions on unfamiliar terrains and the resulting model uncertainty can be used to identify these OOD samples automatically~\citep{beluch2018power, gal2016dropout_uncert, ruuvzivcka2020deep}.
This finding suggests that, in addition to ensembling, future work should focus on curating a more exhaustive set of negative training samples to further improve model robustness.

Our models have been successfully integrated into the operational workflow of the MARS at UNEP's IMEO. Over approximately \fix{\howmanymonthsdeployedN}months of deployment, the system has facilitated the verification of \fix{\totalleaks} methane leak events, leading to \fix{\totalnotified} formal stakeholder notifications and \fix{at least one methane leak mitigation made possible by machine learning} (shown as first row in Figure \ref{fig:mitigation_cases}). Crucially, the system enables the discovery of previously unknown emission sources by systematically monitoring coverage of MARS and enhances the comprehensiveness of global methane emission observations.

Looking ahead, we identify two promising research directions for further reducing false positive detections.
One approach is to implement a post-processing filter based on the spectral similarity of model-delineated plumes, a technique explored in~\citep{xiang2025identification}.
A more fundamental alternative involves developing end-to-end machine learning models that learn directly from radiance data~\citep{HyperspectralViTs}.
Such an approach could yield significant performance gains by bypassing the reliance on intermediate methane enhancement products, which can introduce noise and limit accuracy.
Another promising research direction is the integration of diverse, expert-annotated datasets to create more nuanced ground truth labels.
Combining the labels from this work with those recently released by~\citep{green2023_JPL_labels}, for instance, would enable a systematic analysis of inter-expert agreement.
This approach mirrors advances in other remote sensing fields, such as cloud segmentation, where multi-class labels have proven superior to simple binary masks~\citep{aybar2022cloudsen12}.
For methane detection, such refined labels would be invaluable, allowing models to differentiate between strong and weak emissions to characterize plume morphology by distinguishing the dense source from its more diffuse tail.

It is crucial to recognize that our models and datasets are part of a living system, continuously updated with newly discovered and labeled events. This operational context, with its daily influx of data, makes it an ideal candidate for advanced machine learning paradigms. Methodologies from the Never Ending Learning (NEL) literature \citep{mitchell2018neverendinglearning}, Deep Active Learning (AL) \citep{ruuvzivcka2020deep}, or Continual Learning (CL)~\citep{continuallearning2024} could be very beneficial for ensuring the system's lifelong, iterative improvement and sustained accuracy and robustness.

In conclusion, the strong generalization capabilities of our models across multiple high spatial resolution imaging spectroscopy missions highlight their immediate and future utility. While performance can be enhanced through fine-tuning or advanced techniques like generative style transfer between instruments~\citep{mancoridis2025methaneCycle}, the robust zero-shot performance observed in our study is particularly significant. This capability is critical for the rapid, effective analysis from new and upcoming satellite missions, paving the way for a truly global, responsive, and AI-driven methane monitoring system-a critical tool for enabling targeted mitigation actions and addressing a key driver of near-term climate change.

\section{Data and code availability}\label{sec:public_release}

\fix{
MARS EMIT, EnMAP and PRISMA datasets are publicly available on: \url{https://huggingface.co/datasets/UNEP-IMEO/MARS-Hyperspectral}. Model weights are kept separate at: \url{https://huggingface.co/UNEP-IMEO/MARS-Hyperspectral-models}. Open source codebase which allows full reproduction of our results and full model training is at: \url{https://github.com/UNEP-IMEO-MARS/marsml-hyperspectral}.
}

\section{Acknowledgements}

We would like to thank the analysts of MARS UNEP IMEO for their valuable contributions in identifying methane plumes. In addition to the authors of this paper, we gratefully acknowledge the efforts of Marc Watine and Adriana Valverde. We acknowledge the rest of the MARS team Meghan Demeter, Tharwat Mokalled, Giulia Bonazzi, Florencia Carreras, Konstantin Kosumov and Queen Safari, for their work in country engagement and in the notification process. A portion of this work was carried out at the Jet Propulsion Laboratory, California Institute of Technology, under a contract with the National Aeronautics and Space Administration (80NM0018D0004). Part of Vít Růžička's research was supported by an appointment to the NASA Postdoctoral Program at the Jet Propulsion Laboratory, administered by Oak Ridge Associated Universities under contract with NASA.

\section{Appendix A. Ablation studies}

\subsection{Machine learning architectures}\label{sec:ablation_architectures}

Table \ref{tab:ablation_architectures} shows an extensive ablation study of different machine learning architectures and encoder models. We explore two following directions. First, for the base architecture choice, in addition to U-Net, we also consider DeepLabV3 \citep{chen2017rethinking_deeplabv3}, DeepLabV3+ \citep{chen2018encoder_deeplabv3plus} and U-Net++ \citep{zhou2019unet_pp}. Second, for the exploration of the encoder network, in addition to the default MobileNetV3, we add ResNet34, ResNet50 \citep{he2016deep_resnet}, Xception \citep{chollet2017xception} and InceptionV4 \citep{szegedy2017inceptionV4}. We shows the results over 5 repeated training runs alongside with the changing number of trainable parameters (from 4.71 to 48.79M).

Interestingly, we observe relatively similar results regardless of the used base architecture or encoder model. There are small variations, for example we see very small improvements when using the U-Net++ over the default U-Net model, these are however within each others standard deviations. The larger model architectures seem to produce somewhat reduced number of False Alarms, but this is at the cost of slightly reduced number of Detected events. This might be interesting to explore in future research with growing dataset sizes.

We highlight here, that these changes are however much smaller than those we observed when chosing between different matched filter products (for example when going from the Mag1c product to the WMF product).

\begin{table*}[]
\caption{Ablation study of machine learning architectures and encoders. For all experiments we keep the same input data configuration (RGB+WMF). For each we show the average of 5 training runs.} \label{tab:ablation_architectures}
\centering
\scalebox{0.75}{
\begin{tabular}{@{}lcllllllll@{}}
\toprule
  &  & \multicolumn{4}{l}{Tiled dataset, segmentation:}                        & \multicolumn{3}{l}{Fulltile sources, individual events:} \\
Architecture & Parameters & AUPRC & F1 & Precision & Recall & Detected & Missed & False Alarms \\ \midrule
U-Net with MobileNetV3 (default) & 6.69M      & 68.26  ±3.06 & 63.07  ±2.46 & 72.99  ±1.17 & 55.65  ±3.82 & 213.0  ±20.0  & 98.0  ±20.0   & 3565.2  ±1829.8  \\ \midrule
DeepLabV3 with MobileNetV3         & 11.02M     & 67.59 ±0.82  & 62.78 ±0.54  & 63.17 ±2.19  & \textbf{62.59 ±2.78}  & 196.20 ±27.37 & 114.80 ±27.37 & 4370.20 ±2541.68 \\
DeepLabV3+ with MobileNetV3        & 4.71M      & 67.66 ±2.60  & 62.83 ±2.00  & 69.59 ±5.05  & 57.80 ±4.75  & 200.00 ±21.23 & 111.00 ±21.23 & 3245.40 ±1629.27 \\
U-Net++ with MobileNetV3           & 6.92M      & \textbf{68.90 ±2.68}  & \textbf{64.37 ±1.73}  & 69.43 ±0.64  & 60.08 ±3.08  & \textbf{222.60 ±24.25} & \textbf{88.40 ±24.25}  & 3856.00 ±1399.95 \\ \midrule
U-Net with ResNet34                & 24.44M     & 66.40 ±1.72  & 61.09 ±1.79  & \textbf{74.65 ±2.69}  & 51.96 ±4.03  & 206.00 ±9.65  & 105.00 ±9.65  & 1922.00 ±707.60  \\
U-Net with ResNet50                & 32.52M     & 66.35 ±0.79  & 62.46 ±1.02  & 73.53 ±2.39  & 54.39 ±2.14  & 205.40 ±9.41  & 105.60 ±9.41  & 1937.60 ±340.40  \\
U-Net with Xception                & 28.77M     & 66.73 ±1.60  & 61.70 ±1.85  & 72.57 ±4.83  & 54.13 ±4.52  & 200.00 ±18.95 & 111.00 ±18.95 & \textbf{1825.20 ±918.65}  \\
U-Net with InceptionV4             & 48.79M     & 66.73 ±1.67  & 61.43 ±2.03  & 73.20 ±3.49 & 53.25 ±4.25  & 197.40 ±23.23 & 113.60 ±23.23 & 2091.40 ±1542.33  \\ \bottomrule
\end{tabular}
}
\end{table*}

\subsection{Model ensembles}\label{sec:ablation_ensembles}

Table \ref{tab:ablation_ensembles} shows an ablation study over the number of models used in the ensemble. 
We start with a family of 12 trained models (all using the same settings and hyperparameters as the main used model - U-NET with MobileNetV3 on RGB+WMF) and repeatedly selected a subset of the desired number of models in the ensemble (for example: ensemble of 2 therefore selects non-repeating pairs of models from this family of available models).

We observe, that with increasing number of models in the ensemble, the scores over tiled datasets raise a little bit (the largest improvement is when we go from a single model to at least two models in the ensemble). More importantly the statistics of the number of False Alarms seem to decrease with larger number of models in the ensemble. We observe that with about 5 models, we obtain good performance and going beyond doesn't bring significant improvements.

We also explored weighed model ensembles. Rather than just averaging the model predictions, we assigned each model in the ensemble a weight from the results obtained on the validation set. We however didn't observe any significant improvements over the basic averaging - this is likely due to the fact that the scores on the validation dataset were very close to each other.

\begin{table*}[]
\caption{Ablation study of model ensembles. We start with a family of 12 trained models, each using the same U-Net RGB+WMF configuration. On each row in this table we show the average performance of 5 randomly sampled subsets of models according to the ensemble size. For the single model scenario, we reuse the result from Table \ref{tab:emit_results} as a simple reference.} \label{tab:ablation_ensembles}
\centering
\scalebox{0.88}{
\begin{tabular}{@{}llllllll@{}}
\toprule
Ensemble & \multicolumn{4}{l}{Tiled dataset, segmentation:}                        & \multicolumn{3}{l}{Fulltile sources, individual events:} \\
size & AUPRC & F1 & Precision & Recall & Detected & Missed & False Alarms \\ \midrule
1 (single models)                                 & 68.26 ±3.06 & 63.07 ±2.46 & 72.99 ±1.17 & 55.65 ±3.82 & 213.0 ±20.0  & 98.0 ±20.0   & 3565.2 ±1829.8  \\ \midrule
2                                 & 71.50±2.10  & 64.75±1.60  & 75.75±0.96  & \textbf{56.58±2.20}  & \textbf{221.20±14.08} & \textbf{89.80±14.08}  & 3089.60±1084.41 \\
3                                 & 72.32±1.40  & 64.98±1.11  & 77.61±1.45  & 55.91±1.46  & 217.40±14.57 & 93.60±14.57  & 2283.60±636.35  \\
4                                 & 71.87±0.63  & 64.67±0.30  & 77.11±1.06  & 55.69±0.55  & 210.80±11.75 & 100.20±11.75 & 2235.20±613.59  \\
5                                 & 72.97±0.58  & 65.27±0.35  & 77.97±1.01  & 56.13±0.71  & 211.40±9.97  & 99.60±9.97   & \textbf{1739.60±196.26}  \\
6                                 & 73.11±0.66  & 65.13±0.62  & 78.16±0.90  & 55.85±1.25  & 217.20±8.35  & 93.80±8.35   & 1793.60±299.75  \\
7                                 & \textbf{73.22±0.53}  & \textbf{65.32±0.24}  & 78.18±0.98  & 56.11±0.70  & 216.60±5.46  & 94.40±5.46   & 1821.40±204.95  \\
8                                 & 73.12±0.54  & 65.15±0.21  & \textbf{78.30±0.64}  & 55.78±0.39  & 217.20±9.62  & 93.80±9.62   & 1804.00±263.09  \\ \bottomrule
\end{tabular}
}
\end{table*}

\section{Appendix B. Detailed operational deployment results}\label{sub_operational}

\begin{figure}[!h]
    \centering
    \includegraphics[width=1.0\linewidth]{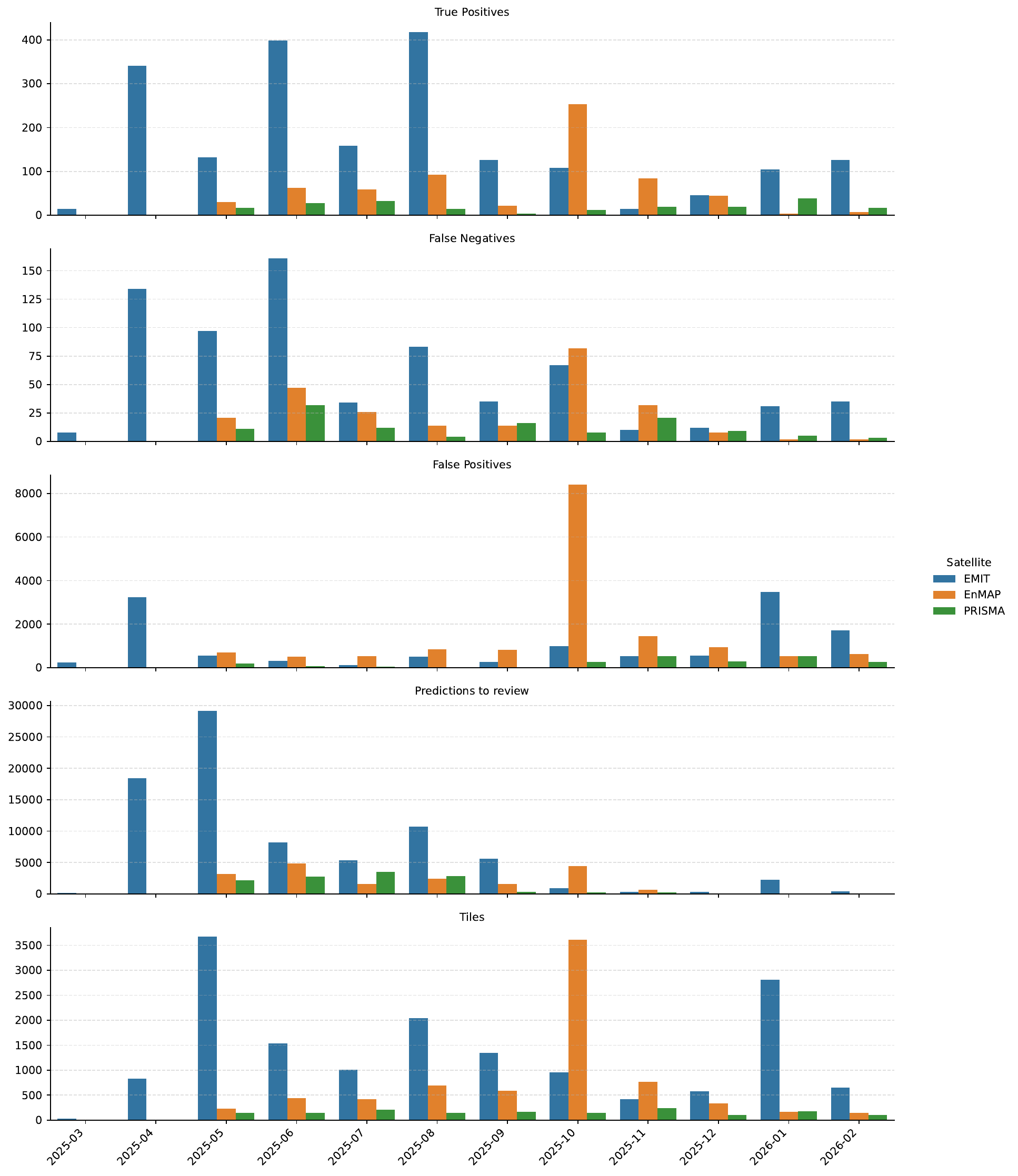}
    \caption{Operational statistics over months of deployment. Number of Tiles refers to the number of full scenes ingested by the MARS team (considering data from hyperspectral satellites: EMIT, EnMAP and PRISMA). True Positives marks successful detections, False Negatives refers to events found only manually by the analysts and False Positives refers to the number of false alarms present on the analyst inspected scenes. Finally, we also show how many alerts were not yet inspected by the analysts (in Predictions to review).}
    \label{fig:deployed_over_months}
\vspace{-2mm}
\end{figure}

Figure \ref{fig:deployed_over_months} shows the operational deployment of the models explored month by month. Among other things, this shows the operational load over the duration of the model's deployment. In April and May 2025, first large wave of ML processed EMIT scenes was added into the system, the models for EnMAP and PRISMA followed from May 2025. These however depended on a manual ingestion of scenes - as such their frequency is much smaller than for EMIT.
In October 2025 an effort was made to ingest larger numbers of EnMAP scenes - this was followed in November 2025 when an automated access via API was added to EnMAP.
Each large data ingestion wave can be also seen in the increased number of detected events and triggered false alarms (as such these are not uniform over months).

In Tables \ref{tab:op_per_sat} and \ref{tab:op_per_sec} we provide a detailed breakdown of predictions by sector (note that sector has been assigned only for plume events and not for all false alarms) and satellite showing the numbers of detected and missed events.
We highlight the lower recall on the coal sector, these events often occur in mountainous regions where plume shapes are significantly different than in oil and gas plain areas.
Additionally, a systematic survey of coal mine emissions was carried out during these months as part of the IMEO's Steel Methane Programme.
In contrast with the Oil and Gas sector, both Coal and Waste are underrepresented in the original training dataset (as can be seen in Figure \ref{fig:emit_stratification}).
The number of predictions to review (these are predictions made by the model, which were not yet validated by the analysts), further demonstrates the need for a more automated prioritization and plume vetting scheme - we believe that methods proposed by \citep{xiang2025identification} could help to automatically reject some of these predictions.

\begin{table}[h]
\caption{Operational deployment results. Per-satellite breakdown}
\label{tab:op_per_sat}
\centering
\scalebox{1.0}{
\begin{tabular}{@{}lcccc@{}}
\toprule
Satellite                                  & {TP} & {FN} & {FP} & Predictions to review \\ \midrule
EMIT      &    1990  &  707	 & 12469   &     81900   \\
EnMAP     & 	659  &  248   & 14066    &    18507   \\
PRISMA    & 	202  & 121     &  2203     &   12371  \\ \bottomrule
\end{tabular}
}
\end{table}

\begin{table}[h]
\caption{Operational deployment results. Per-sector breakdown}
\label{tab:op_per_sec}
\centering
\scalebox{1.0}{
\begin{tabular}{@{}lccc@{}}
\toprule
Sector      & 		TP   	&	FN  	&	Recall  \\  \midrule
Oil and Gas   & 		1252   & 		331    & 		79.1  \\
Coal          & 		1005  	 & 	589   	 & 	63.0   \\
Waste        	 & 	577  	 & 	135    & 		81.0  \\
Other           	 & 	17    & 		21  	 &  	44.7  \\ \bottomrule
\end{tabular}
}
\end{table}

\begin{figure*}[h]
    \centering
    \includegraphics[width=0.48\textwidth]{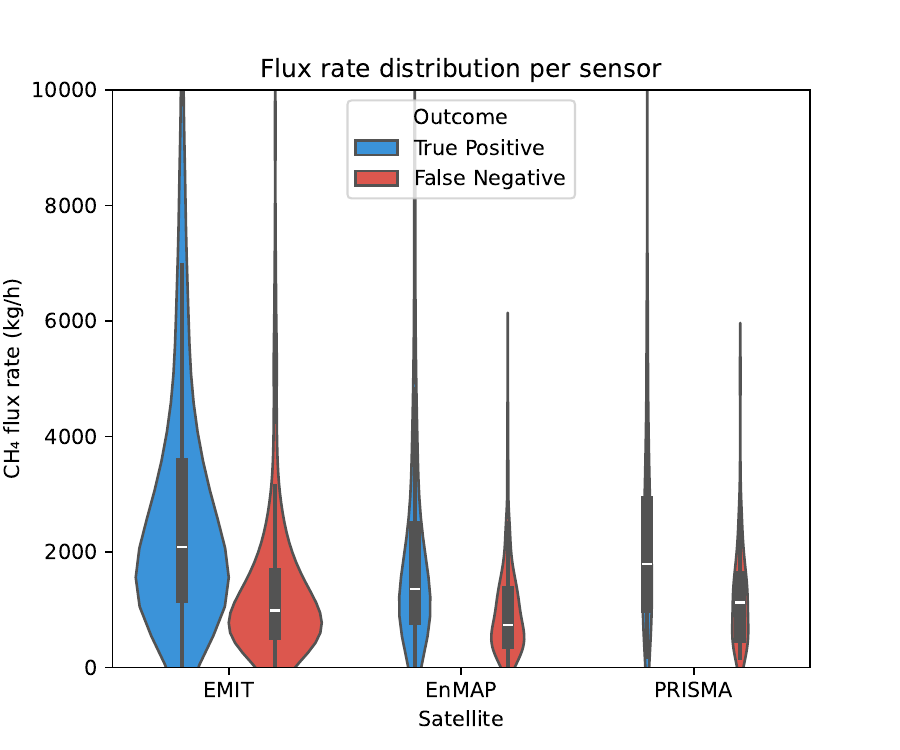}
    \includegraphics[width=0.48\textwidth]{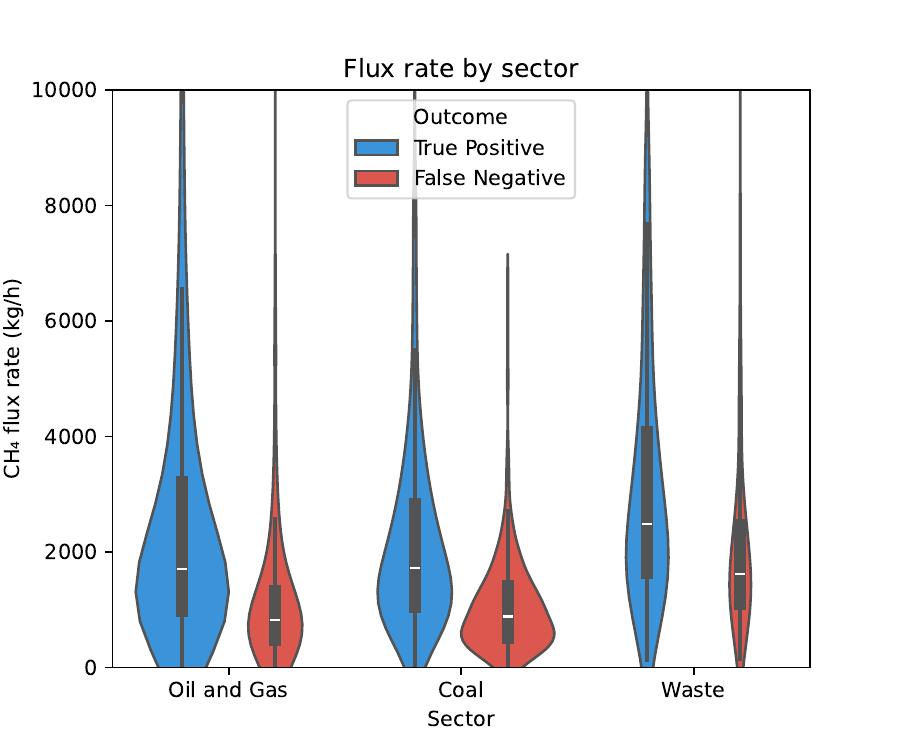}
    \caption{Flux rate of detected and missed events separated by sensor and sector. Width of each plot represents the total number of plumes.}
    \label{fig:flux_by_sensor_sector}
\end{figure*}

Finally, in Figure \ref{fig:flux_by_sensor_sector} we explore the flux rate distribution of the detected and missed events. 
In all variants, we see a separation between the model detected events and those that were missed. Missed events tend to have lower flux rate.
We don't observe large differences across sectors, except that Waste tends to have slightly higher flux rates on average.

\section{Appendix C. Extended datasets release}\label{sup_extended_dataset}

We are releasing an extended version of the dataset used in the main results of this paper. Since the deployment of our models, the operational archive used by UNEP IMEO's MARS team has significantly grown. More months were added - the original dataset ended with few events in March 2025, the extended variant contains events until the end of December 2025. Additionally, more methane plumes were found in the archives of EMIT, EnMAP and PRISMA data.
Table \ref{tab:dataset_statistics} shows the detailed statistics of the extended datasets.
We call this extended version the v2025 data, and it is hosted alongside the first released version on: \url{https://huggingface.co/collections/UNEP-IMEO/mars-hyperspectral}. 
Both datasets are formatted in the same manner, and provided with splits following the same methodology (temporal division in EMIT data and spatial in EnMAP and PRISMA). For EMIT data, the same months are kept for Validation and Test susets, as such it is possible to compare the trained models accross these datasets (but keep in mind that the number of events in these subsets is much larger in the v2025 release).
We believe that the public release of this dataset will help research in this field.

\begin{table}[h]
\caption{Detailed statistics of the extended variants of datasets.}
\label{tab:dataset_statistics}
\centering
\scalebox{1.0}{
\begin{tabular}{@{}lccc@{}}
\toprule
\multicolumn{4}{l}{\textbf{MARS-EMIT v2025 dataset}}                                                       \\ \midrule
\textbf{Split} & \textbf{Pos. : Neg. tiles} & \textbf{Sites}       & \textbf{Original granules} \\
Train          & 5743 : 7112              & 3460                 & 4742                       \\
Val            & 1729 : 2422               & 1982                 & 1422                        \\
Test           & 1322 : 1725                & 1767                 & 1089                        \\ \midrule
               & \multicolumn{1}{l}{}            & \multicolumn{1}{l}{} & \multicolumn{1}{l}{}       \\ \midrule
\multicolumn{4}{l}{\textbf{MARS-PRISMA v2025 dataset}}                                                     \\ \midrule
\textbf{Split} & \textbf{Pos. : Neg. tiles} & \textbf{Sites}       & \textbf{Original granules} \\
Train          & 871 : 1939                 & 1184                  & 765                        \\
Test           & 292 : 409                 & 483                  & 257                        \\ \midrule
               & \multicolumn{1}{l}{}            & \multicolumn{1}{l}{} & \multicolumn{1}{l}{}       \\ \midrule
\multicolumn{4}{l}{\textbf{MARS-EnMAP v2025 dataset}}                                                      \\ \midrule
\textbf{Split} & \textbf{Pos. : Neg. tiles} & \textbf{Sites}       & \textbf{Original granules} \\
Train          & 926 : 1354                 & 1249                  & 891                        \\
Test           & 338 : 594                   & 685                  & 437                         \\ \midrule
\end{tabular}
}
\end{table}

\bibliography{bib}

\end{document}